\pdfoutput=1

\documentclass[10pt,article,compsoc]{IEEEtran}

\usepackage[hidelinks]{hyperref}
\ifCLASSOPTIONcompsoc
  \usepackage[nocompress]{cite}
\else
  \usepackage{cite}
\fi

\usepackage{bbm}

\ifCLASSINFOpdf
  \usepackage[pdftex]{graphicx}
\else
\fi

\begin{document}
\title{Stroke-based Rendering:\\From Heuristics to Deep Learning}

\author{Florian~Nolte,
        Andrew~Melnik,
        and~Helge~Ritter%
\IEEEcompsocitemizethanks{
\IEEEcompsocthanksitem F. Nolte, A. Melnik and H. Ritter are with the Center for Cognitive Interaction Technology at Bielefeld University, Germany.
}

\protect%
E-mail: \ \{florian.nolte, andrew.melnik\}.papers@gmail.com\\ %
\url{https://github.com/ndrwmlnk/stroke-rendering}
}

\IEEEtitleabstractindextext{%
\begin{abstract}
In the last few years, artistic image-making with deep learning models has gained a considerable amount of traction. A large number of these models operate directly in the pixel space and generate raster images. This is however not how most humans would produce artworks, for example, by planning a sequence of shapes and strokes to draw. Recent developments in deep learning methods help to bridge the gap between stroke-based paintings and pixel photo generation. With this survey, we aim to provide a structured introduction and understanding of common challenges and approaches in stroke-based rendering algorithms. These algorithms range from simple rule-based heuristics to stroke optimization and deep reinforcement agents, trained to paint images with differentiable vector graphics and neural rendering.

\end{abstract}

}

\maketitle

\IEEEdisplaynontitleabstractindextext

\IEEEpeerreviewmaketitle

\IEEEraisesectionheading{\section{Introduction}\label{sec:introduction}}

\IEEEPARstart{I}{n} recent years, many machine learning and deep learning models have been successfully developed for the purpose of creating unique and artistic images \cite{karras2021alias, ramesh2022hierarchical, crowson2022vqgan}.
Style-transfer models especially are frequently used to replicate photographs in the style of reference paintings \cite{gatys2016image, huang2017arbitrary}. Typically, these image generation models directly predict the pixel values of raster images
through operations such as convolutions \cite{goodfellow2014generative, kingma2013auto}. This is however very different from the way humans generate and perceive images. Instead of pixel values, we think and work with simple shapes and strokes. Teaching machines to paint like humans might provide useful insight into our artistic process and decision-making \cite{fernando2021generative}.

\begin{figure}[!t]
\centering
\includegraphics[width=3.4in]{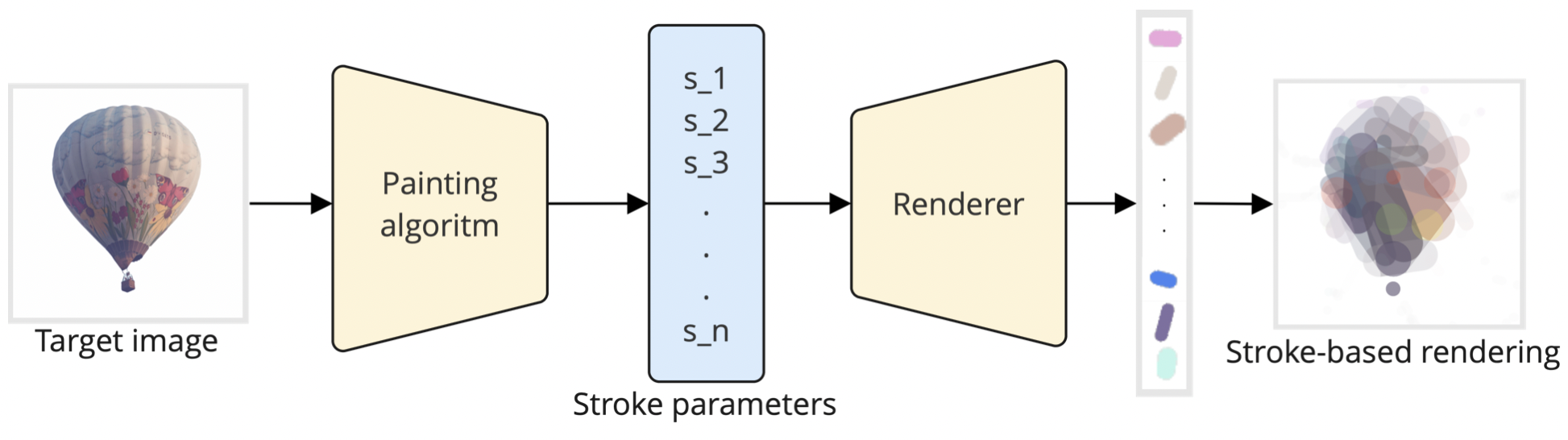}
\caption{In stroke-based painting algorithms, a target (e.g. image reference) is processed through a painting algorithm. This algorithm predicts parameters of shapes, which can be rasterized through a rendering engine.}
\label{fig:painting_algorithms}
\end{figure}

The paradigm of stroke-based rendering (SBR) was first introduced through the seminal work of Paul Haeberli \cite{haeberli}. It mimics the human painting style by generating structured images that are made up of parameterized brushstrokes (Figure \ref{fig:painting_algorithms}), similar to vector graphics. Depending on the algorithm, it is possible to do stroke-based style-transfer \cite{zou2021stylized, kotovenko2021rethinking} (Figure \ref{fig:style_transfer}), paint videos \cite{hertzmann2000} (section \ref{video_painting}) and visualize text \cite{frans2021clipdraw} as well as semantic categories of objects \cite{ganin2018synthesizing}.

There are several potential benefits of constructing stylized images from a sequence of overlapping strokes and shapes. By using brushstrokes to draw an image instead of imitating a painted look with pixel operations \cite{melnik2022faces}\cite{melnik2022face}, a hand-painted aesthetic is easily accomplished, while preserving desirable image properties of vector graphics \cite{li2020differentiable}, such as easy scaling to high resolutions and low file sizes. Furthermore, the painted images are easily editable for artists, fit well into digital drawing programs \cite{xie2014portraitsketch}, and the appearance of the final image can be intuitively specified through the stroke model and the hyperparameters of the painting algorithms \cite{kang2006unified}.

The main goal and contribution of this survey is, to give a comprehensive overview of past and current painterly rendering algorithms. For this, we categorize the algorithms by the machine learning and computer graphics techniques they use to decide on the stroke parameters. In addition to high-level descriptions of the different approaches, we provide a taxonomy and database of the painting algorithms. We focus on understanding and comparing which image features and methods are appropriate for deciding on brushstroke parameters, but try to avoid discussing other implementation details of the algorithms. We only include algorithms published before 2022 that place sequences of simple, colored shapes or strokes, in order to create a digital image with a painted look. In addition, we present a short overview of techniques used for generating stroke-based videos and animation. Through this survey, we hope to give inspiration for the development of new rendering algorithms that utilize not just the latest developments from deep learning, but also take older approaches into consideration and maybe even reinvent them to fit into new learning-based architectures. Unlike other surveys \cite{kyprianidis2012state, hertzmann2003survey, xu2022deep, kumar2019comprehensive}, we provide a structured introduction of stroke-based painterly rendering instead of the broader topic of non-photorealistic rendering and discuss the design choices and goals of a large number of algorithms.

\section{Related work}

Stroke-based rendering is an example of non-photorealistic rendering (NPR) \cite{NPAR2000} which emerged as a sub-field of image processing and computer graphics and is concerned with generating artistic and stylized images instead of photorealistic ones\cite{kumar2019comprehensive}. NPR consists of stroke-based algorithms, as well as pixel-based techniques.

\subsection{Pixel-based image generation} 
Examples of pixel-based NPR are image analogies for style transfer\cite{hertzmann2001image, gatys2016image}, region-based abstraction \cite{decarlo2002stylization} and image filtering \cite{kumar2019comprehensive}. Lately, deep learning methods have conquered the field of NPR, which able to generate almost arbitrary artistic and realistic images with minimal human supervision \cite{goodfellow2016deep}. There exist a large number of possible architectures for generating images with neural networks, such as GANs \cite{goodfellow2016deep}, variational autoencoders \cite{kingma2013auto} and diffusion models \cite{dhariwal2021diffusion}. Typically, these use pixel-based operations such as transposed convolutions \cite{dumoulin2016guide} to accomplish NPR and computer vision tasks like generating images \cite{goodfellow2014generative}, visualizing text prompts \cite{crowson2022vqgan, ramesh2022hierarchical} and changing the style of an image \cite{gatys2016image, huang2017arbitrary}.

\subsection{Stroke-based rendering}

Stroke-based approaches do not just consist of painterly rendering algorithms, but also include other techniques with slightly different goals. Sketch-based image synthesis models \cite{xu2022deep} are closely related to painterly algorithms, but use a small number of non-colored strokes to generate sketches and doodles. For these, there are large vector-based datasets of simple doodles available to train deep learning models \cite{ha2017neural}. Hatching \cite{winkenbach1994computer} and stippling\cite{deussen2000floating} algorithms represent grayscale images through simple dots and lines. Image vectorization \cite{sun2007image, lai2009automatic} does not restrict the output to strokes, but allows more complex shapes. The goal here is usually not an artistic abstraction of images, but a translation between vector and raster graphics. Robotic painting \cite{tresset2013portrait} uses robots to paint and draw with real brushes and pencils instead of digitally simulated paint. Image mosaicking algorithms \cite{hausner2001simulating} place non-overlapping rigid shapes on a canvas, to create a mosaic effect. Stroke-based renderings of 3D models \cite{meier1996painterly} uses additional information about depth and geometry of image contents to guide stroke parameters.

\section{Painting algorithms}
In the following, we provide an overview of different painterly rendering algorithms and their building blocks, sorted by our taxonomy (section \ref{taxonomy}). There are a large number of useful techniques, from computer graphics to machine learning, that can guide the parameters of strokes. Algorithms are naturally not limited to using just one method from the image processing toolbox, and they might fall into multiple categories at once. In the following, the different painting approaches are sorted by their most prominent algorithmic choices.

\subsection{Stroke models}
\label{stroke_definition}

\begin{figure}[!t]
\centering
\includegraphics[width=3.4in]{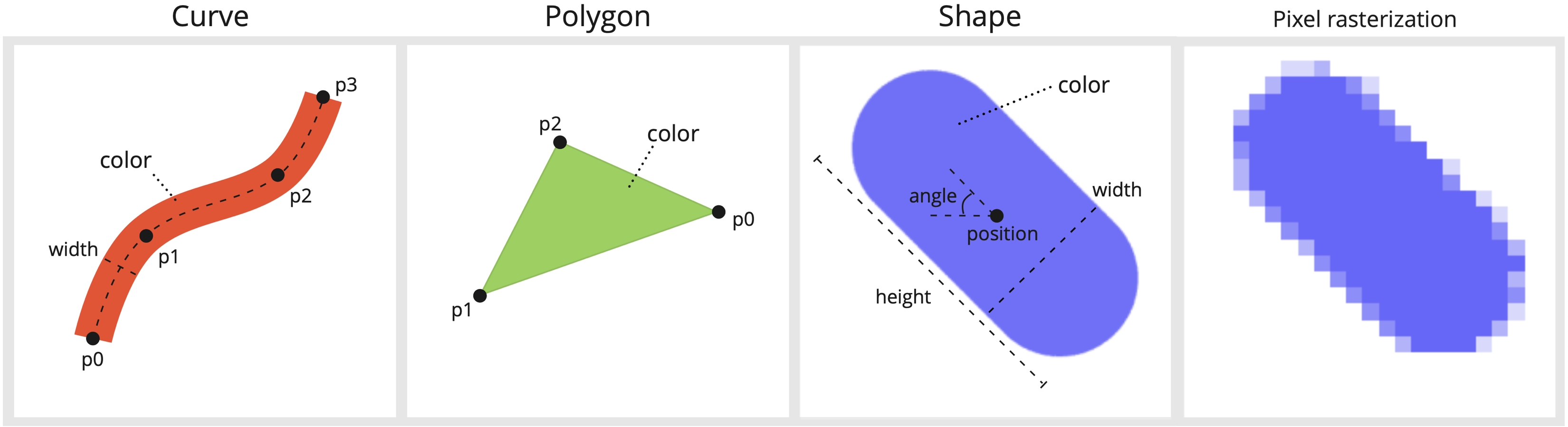}
\caption{There are many different possibilities to choose the parameters and appearance of strokes. Once stroke parameters have been chosen, the vector shapes can be rasterized into an arbitrarily sized pixel image.}
\label{fig:stroke_model}
\end{figure}

SBR algorithms predict stroke parameters to construct an image. However, there are many different possibilities for defining the appearance and parameters of strokes (Figure \ref{fig:stroke_model}). This can have a big impact on the look of the final painting (Figure \ref{fig:brushes_paintings}) and even on the performance of the algorithm. In the following, we give an overview of the most common stroke models, used in painting algorithms (Figure  \ref{fig:neural_painting}). Many painting algorithms use pixel-based textures (Figure  \ref{fig:neural_painting}) to achieve a natural, painterly look for their strokes \cite{hertzmann2002fast}.
Uniformly colored strokes are most common, but multicolored ones are possible as well \cite{zou2021stylized, huang2011painterly}. Apart from the simple strokes covered here, it is of course possible to use more complex simulated painting mediums such as watercolor \cite{curtis1997computer} or ink wash \cite{ning2011contour, xie2013artist} to generate digital paintings.

\textbf{Geometric shapes} are simple strokes, such as ellipses, rectangles or straight lines. Position, angle, width, and height can be changed, and they can be easily textured by replacing the fill color with a brush image. 

\textbf{Curved lines} are usually based on splines or B\'{e}zier curves and enable long, expressive strokes. The position of every control point can be changed, and sometimes additional parameters for varying the stroke width and curvature are used. 

\textbf{Polygons} are shapes made up of multiple connected control points. Depending on the number of points, single polygons can produce complex shapes. 

\subsection{Differentiable rendering}
\label{differentiable_rendering}

\begin{figure}[!t]
\centering
\includegraphics[width=3.4in]{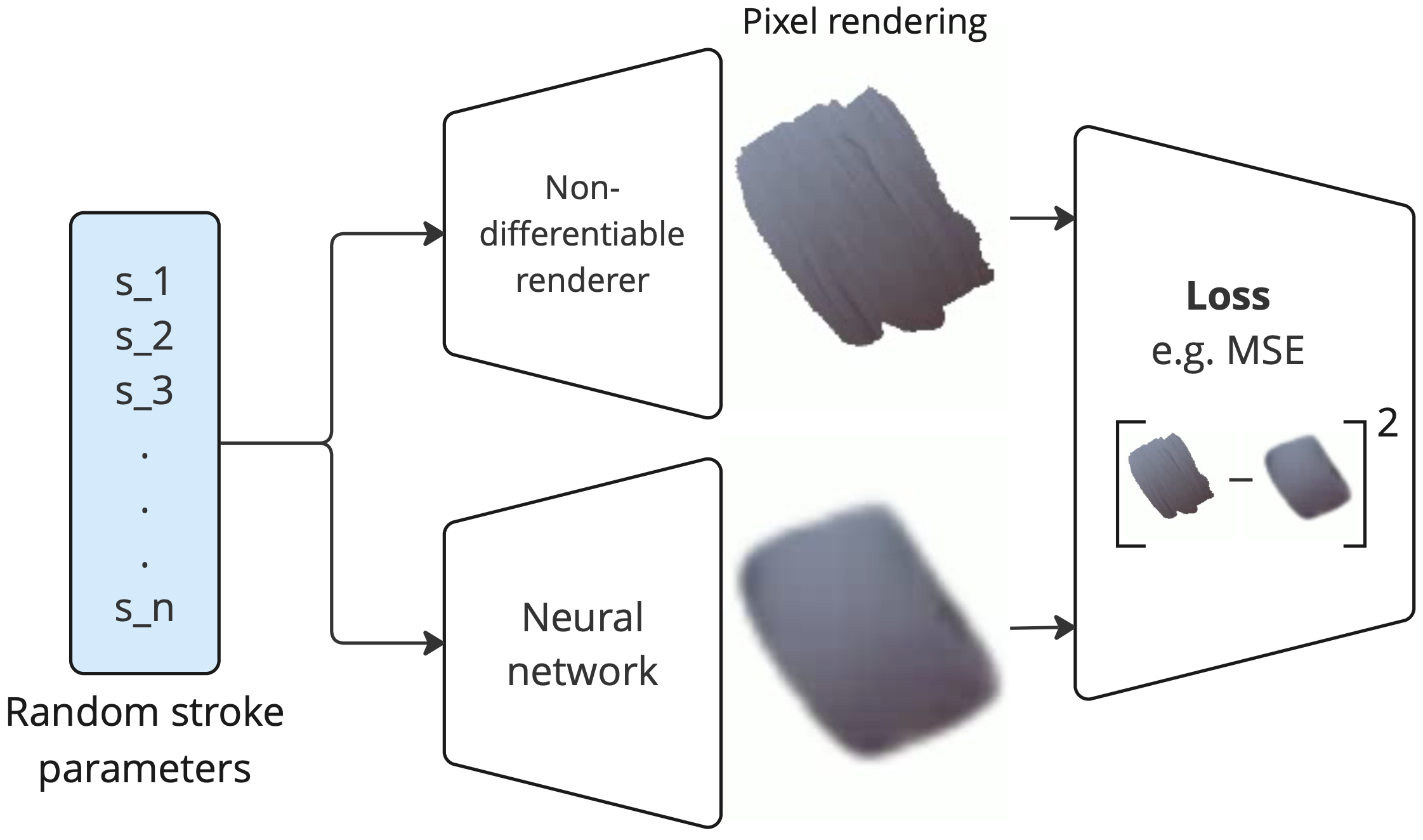}
\caption{A neural renderer can be trained to imitate non-differentiable rendering engines. Different neural architectures and losses can be used during training \cite{nakano2019neural}.}
\label{fig:neural_renderer_training}
\end{figure}

All stroke models from the previous section are defined by their parameters. But in order to visualize a stroke from a given model, it needs to be converted from the parameterized representation into a pixel image (Figure \ref{fig:stroke_model}). This process is carried out through a rendering pipeline, which produces pixel images from sequences of parameters. We restrict the following discussion to 2D rendering of shapes and strokes.

Renderers can be defined as functions $R(S) = I$ that map parameters $S$ to pixel images $I$ through the use of rasterization. In typical rendering pipelines, this rasterization step is either non-differentiable, or has the gradient $\frac{\partial I}{\partial S} = 0$ almost everywhere \cite{mihai2021differentiable}. While many painting algorithms work just fine with traditional, non-differentiable rendering, more recent approaches can greatly benefit from gradient information (see sections \ref{gradient_optim} and \ref{DL_painting} for details). This enables end-to-end differentiable machine learning algorithms that know how they have to change and optimize stroke parameters through gradient descent and backpropagation \cite{mihai2021differentiable}. In essence, differentiable rendering gives models access to fine-grained information about how changes to stroke parameters will affect the pixels of the rendered image.

\begin{figure}[!t]
\centering
\includegraphics[width = 3.3in]{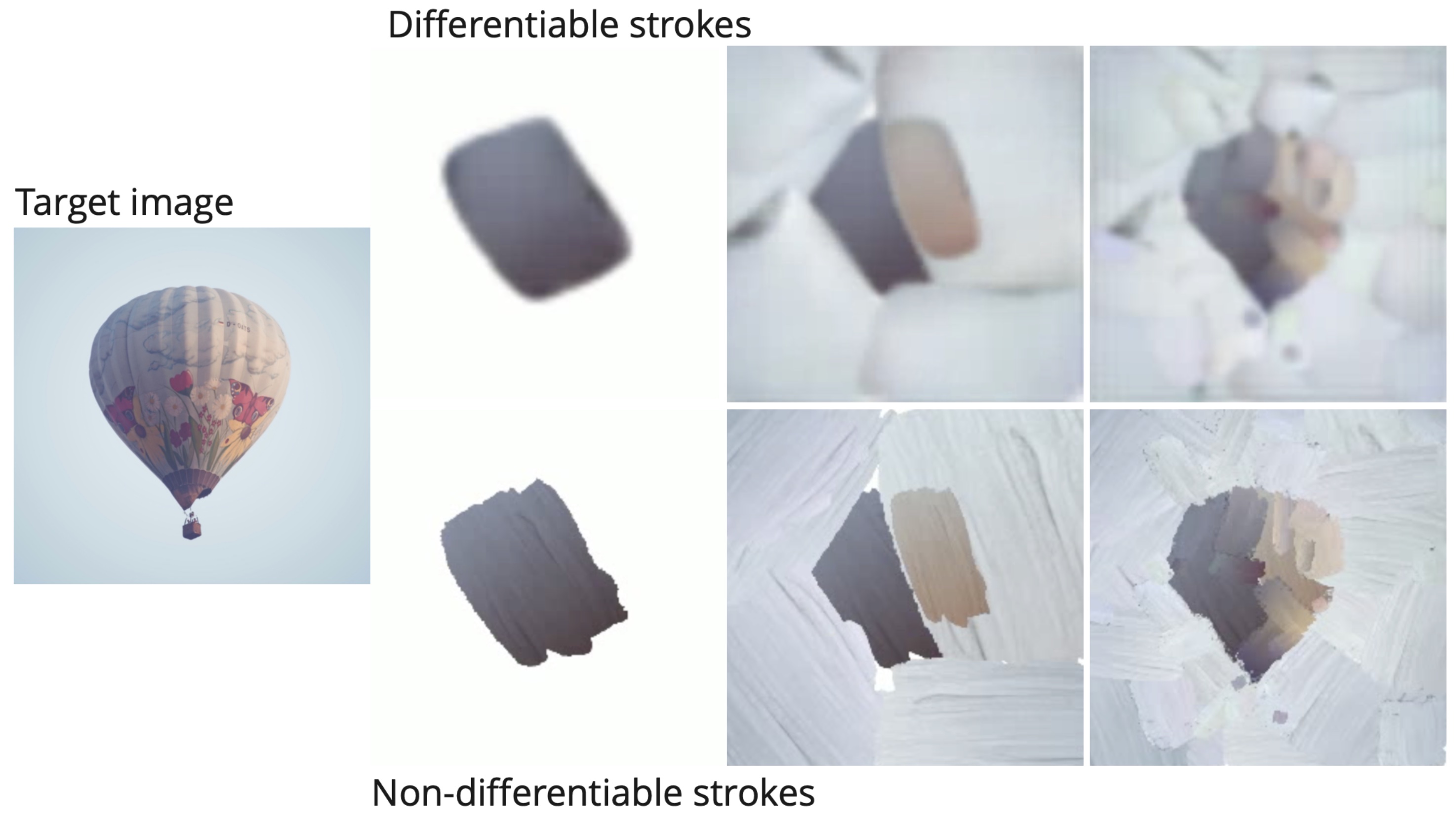}
\caption{Painting process of an SBR algorithm \cite{zou2021stylized} that uses a neural rendering approximation (top row) to calculate parameters of strokes. The strokes are placed one after another on the canvas (left to right). Once all stroke parameters have been calculated, the neural approximation can be replaced by the ground truth strokes (bottom). This is useful if the neural renderer is not able to capture all details of the underlying stroke model.}
\label{fig:neural_painting}
\end{figure}

While the technical details are beyond the scope of this survey, a differentiable renderer can be either hand-crafted or learned through a neural network. Hand-crafted differentiable renderers only use differentiable operations when converting parameters to pixel images \cite{li2020differentiable, mihai2021differentiable, kotovenko2021rethinking, liu2021paint}. Additionally, they make an effort to provide useful gradients, for example by smoothing edges through soft rasterizers \cite{mihai2021differentiable}. Neural renderers are neural networks, trained to approximate a traditional non-differentiable renderer \cite{zou2021stylized, nakano2019neural, huang2019learning} (Figure \ref{fig:neural_renderer_training}). The required dataset of stroke parameters and corresponding pixel images can be efficiently created through a non-differentiable renderer. The performance of the neural renderer depends on the model architecture \cite{nakano2019neural}, and the resulting strokes might not always look perfect (Figure \ref{fig:neural_painting}). Because of this, the strokes from the differentiable renderer are sometimes replaced with the ground truth stroke for the final image, once gradient computations are no longer needed \cite{zou2021stylized}.

\subsection{Taxonomy}
\label{taxonomy}

\begin{figure}[!t]
\centering
\includegraphics[width=3.4in]{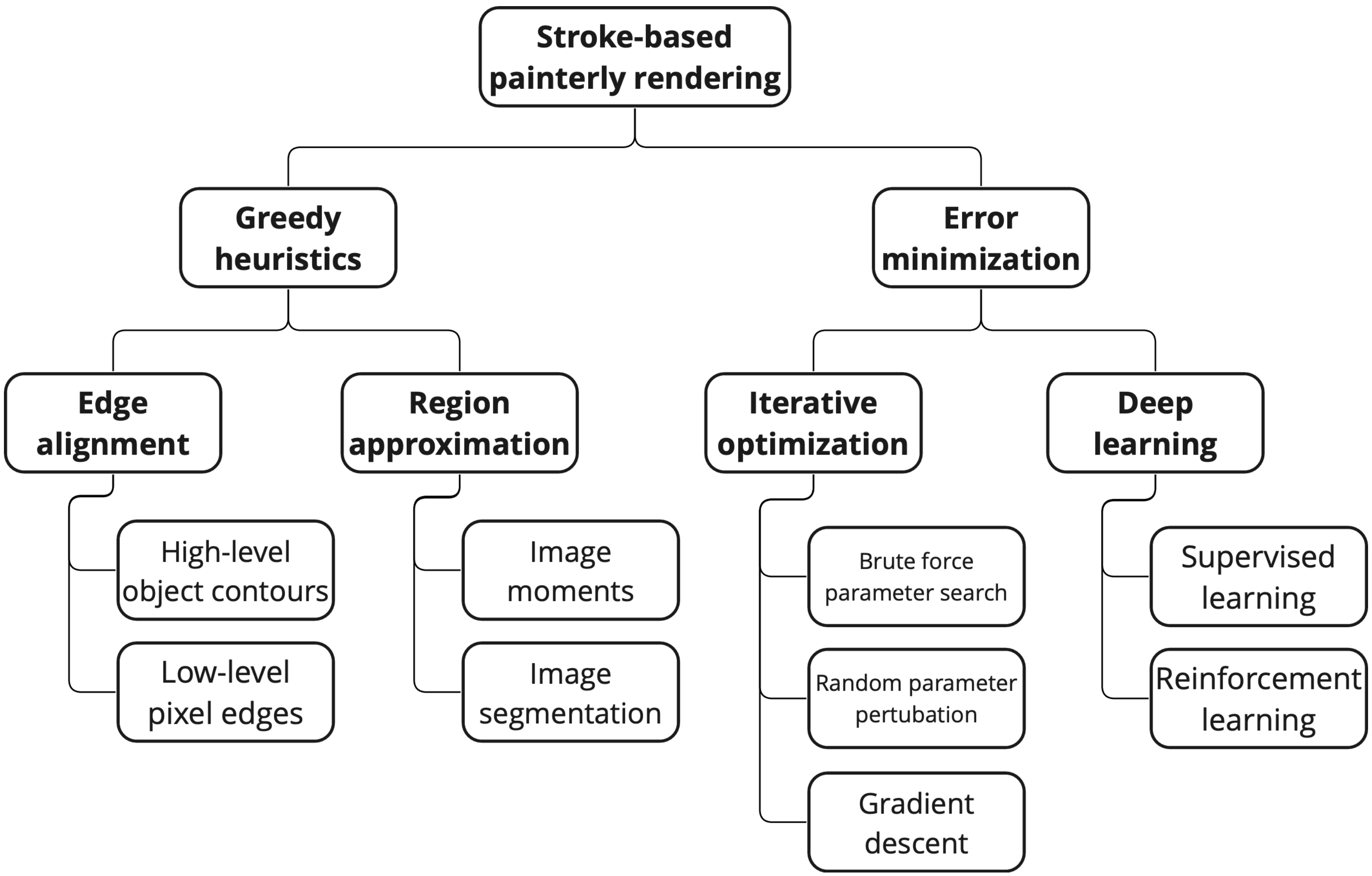}
\caption{Taxonomy of stroke-based painterly rendering algorithms.}
\label{fig:taxonomy}
\end{figure}

We propose a taxonomy of painting algorithms that aims at categorizing them by how they make decisions about the values of stroke parameters (Figure \ref{fig:taxonomy}).

Haeberli \cite{haeberli} not only proposed the concept of SBR but also provided two perspectives on solving the problem. SBR can be approached ``greedily" \cite{hertzmann2003survey}, or from a mathematical optimization perspective.
Within greedy algorithms (section \ref{greedy}), edge alignment approaches (section \ref{edge_alignment}) try to retain low-level pixel edges and high-level object contours in the painting. 
Region approximation algorithms (section \ref{region_approximation}) use image segmentation and other statistical image properties to calculate the shape and size of the strokes. Error minimization (section \ref{error_minimization}) can be divided into iterative optimization algorithms and deep learning. In iterative algorithms (section \ref{iterative_optim}), stroke parameters are repeatedly changed according to brute-force, random, or gradient-based strategies in order to minimize an image error. Deep learning approaches (section \ref{DL_painting}) try to learn optimal painting policies with supervised or reinforcement learning. %

The classification into greedy and optimization approaches is in line with the taxonomy of Hertzmann \cite{hertzmann2003survey}, although there are other ways to group the algorithms. For example, the methods can be categorized according to whether they use the entire image or only local regions of it to calculate the stroke parameters \cite{kyprianidis2012state}. Greedy algorithms are usually local and optimization approaches are often global methods. Additionally, algorithms can be sorted according to whether they require any human input and by their use of low-level and high-level image features. %

\subsection{Greedy Approaches}
\label{greedy}

Greedy algorithms use hand-crafted rules in a bottom-up painting approach. Stroke parameters are directly calculated in a single pass from the target image. They use image processing techniques such as edge detection and segmentation to place strokes that follow the content of the image. The style of the paintings is largely decided by the algorithm and its hyperparameters. For example, impressionist paintings can be achieved by placing a large number of small strokes (Figure \ref{fig:edge_alignment}) while more abstract paintings might benefit from an algorithm that chooses larger shapes (Figure \ref{fig:segmentation}).

There are two complementary approaches to place strokes that respect the structure of the image. Region-based approaches (section \ref{region_approximation}) try to find homogeneous areas in the image and place strokes on these areas. Edge-based approaches (section \ref{edge_alignment}) align the stroke directions with the edges and gradients of the image. In essence, these both accomplish the same goal: strokes are not simply painted over object boundaries but are placed to match the contents of the image.

\subsubsection{Edge Alignment}
\label{edge_alignment}

\begin{figure}[!t]
\centering
\includegraphics[width=3.5in]{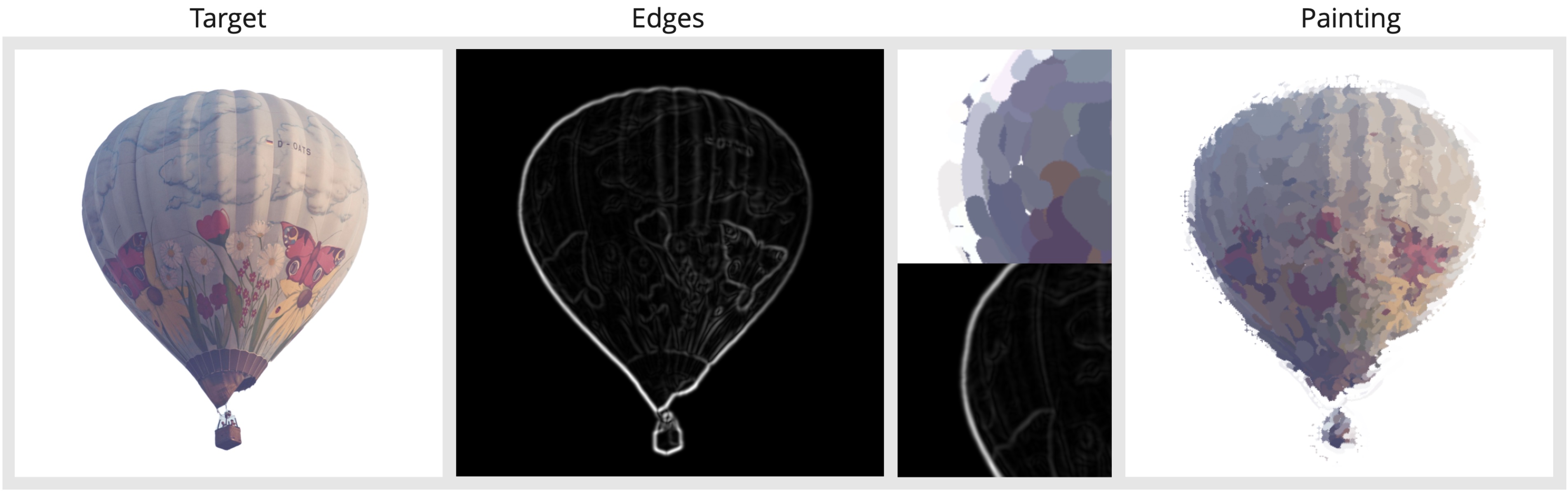}
\caption{Edge alignment painting algorithm \cite{hertzmann1998}, implemented by \cite{schaldenbrandGithub}. Strokes are curved according to the edges of the target, resulting in an impressionist painterly image. The zoomed-in section was created with larger strokes and small color jitter to emphasize the curvature of the strokes.} 
\label{fig:edge_alignment}
\end{figure}

\begin{figure}[!t]
\centering
\includegraphics[width=3.5in]{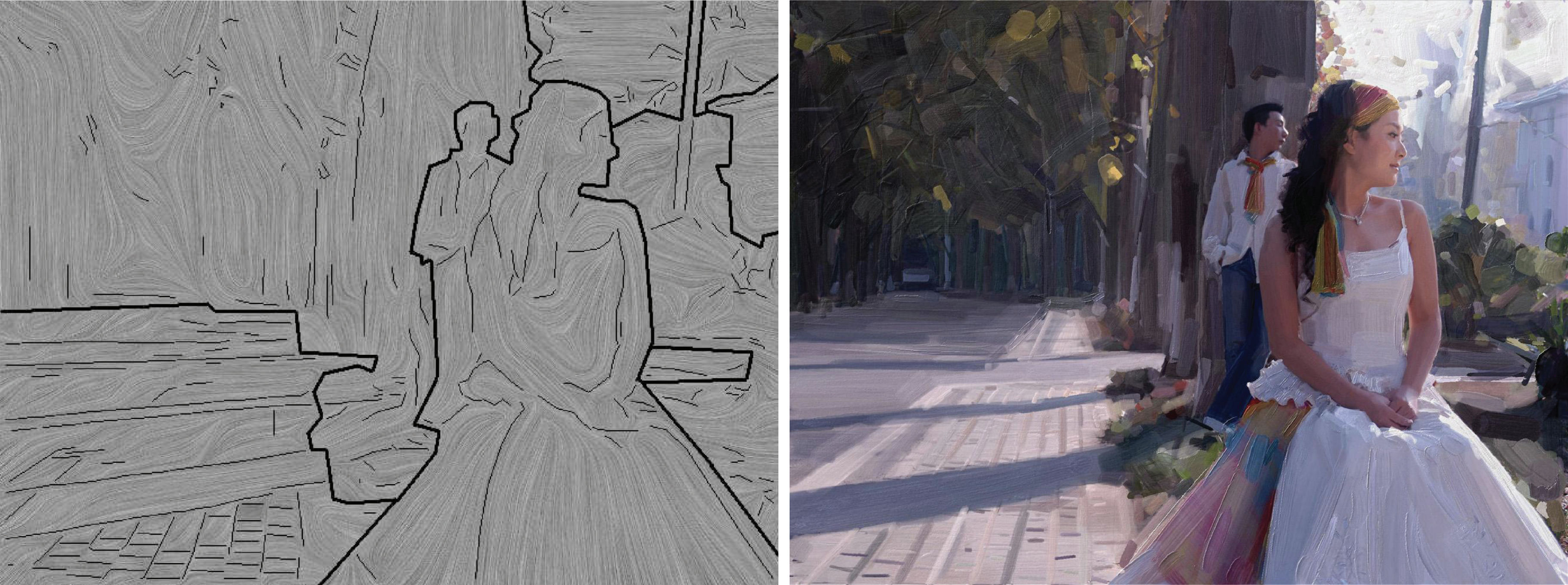}
\caption{Semantic segmentation helps to paint foreground subjects with higher amounts of detail, compared to the background. Stroke directions are aligned with an orientation map, interpolated from  high-level contours and edges (left). Note how the strokes for the dress and the sidewalk follow the image structure. The strokes representing the leaves are more chaotic due to the more complex edges. Figures used with permission from \cite{zeng2009image}.} 
\label{fig:image_parsing}
\end{figure}

Following Haeberli's work \cite{haeberli}, the overall recipe for the following algorithms is mostly the same. Strokes are semi-randomly distributed on the canvas, often with a higher number of strokes being placed in areas with strong edges and important details. The stroke color is chosen by sampling and averaging the pixel colors of the target image at the respective stroke positions. Sometimes, the color and position parameters are slightly jittered to obtain a more natural looking painting \cite{litwinowicz}. Size and orientation are based on the gradients and edges, so that
the strokes follow the structure of the image (see Figure \ref{fig:edge_alignment}).
If the algorithm handles differently sized strokes, large ones are almost always drawn in the background, and small ones on top. Many approaches paint in multiple stacked layers of strokes with decreasing sizes, similar to how humans sometimes paint rough shapes and outlines first and small details in the end. However, different orderings of strokes are possible and will have different effects \cite{northam2010brush}.

There are two types of edges that can be used for stroke alignment: low-level and high-level ones.

Low-level edges signify a strong value change of neighboring pixels and can be calculated using Sobel filtering, the Canny edge detector, or other methods \cite{szeliski2010computer} (Figure \ref{fig:edge_alignment}). A high gradient value represents a strong edge, which is oriented perpendicular to the direction of the steepest ascent in the image \cite{szeliski2010computer}. Therefore, image gradients can be used to calculate edge orientation and strength. Smooth areas in the image do not contain useful edge information, regardless of the edge detection method. Because of this, it is common practice to interpolate the edge orientations of the strongest edges in the image through a variety of techniques \cite{turk1996image, zhang2006interactive, bors2001introduction, franke1979critical} (similar to interpolation in Figure \ref{fig:image_parsing}), rather than using noisy values of low-gradient regions. These interpolated orientation maps can be used to guide the stroke directions of the whole painting based on the most prominent edges in the image. Different strengths of Gaussian blurring are used before the edge detection to remove noise \cite{szeliski2010computer} and emphasize differently sized image details.

High-level edges are more complex boundaries in the image, for example, the actual contours of objects \cite{zeng2009image, kagaya2010video, zhao2010sisley} (Figure \ref{fig:image_parsing}). These are not as easily detectable as low-level edges and often need to be supplied by the user. The object edges can be used in combination with low-level edges to make semantic objects in the image more easily identifiable. Additionally, semantic information about the image content can be used to vary the painting style between different objects. Salience is another high-level image feature which assigns importance values to regions.

\smallskip

\textbf{Low-level edges}

\smallskip

\textbf{``Paint by numbers: Abstract image representations” \cite{haeberli}} introduce the first stroke-based painterly rendering algorithm. Simple strokes are interactively positioned and either rotated by the user or aligned with the nearest image edges. 

\textbf{``Processing images and video for an impressionist effect” \cite{litwinowicz}} generate impressionistic images through the use of a large number of similarly sized strokes. These strokes are positioned on a regular grid and colored according to the pixel colors of the target image. Their orientation and length is chosen to preserve the original edges and gradients in the image. The stroke order and other parameters are then subtly perturbed to create a more natural, painterly look.

\textbf{``Painterly rendering with curved brush strokes of multiple sizes” \cite{hertzmann1998}} are the first to propose an algorithm capable of handling long, curved strokes of different sizes. The strokes are based on cubic B-splines and generated one control point after the other in the direction orthogonal to the local image gradient (Figure \ref{fig:edge_alignment}). Painting happens in multiple layers where larger strokes are drawn according to large-scale gradients, and thin strokes are greedily placed on top of areas where the large strokes are not sufficient to capture all the details. 

\textbf{``Image and video based painterly animation” \cite{hays2004}} use a similar layering strategy to \cite{hertzmann1998}, but with simple, straight strokes, and they make sure to always place small strokes near regions with strong edges. Additionally, they do not use full image gradients but interpolate the direction from the strongest few edges to globally guide the direction of all strokes.

\textbf{``Painterly rendering controlled by multiscale image features” \cite{kovacs2004painterly}} employ a similar layered painting approach to others \cite{hertzmann1998, hays2004}, but utilize a multi-scale edge and ridge map to place strokes.

\textbf{``Interactive painterly stylization of images, videos and 3D animations” \cite{lu2010interactive}} speed up the painting process by enabling the calculation of thousands of stroke parameters in real time on the GPU. For this, the parameters of the strokes have to be independently calculated in parallel and can only be based on information from the local pixel neighborhood of the stroke. They place, size, and orient strokes in multiple layers according to the gradients of the image.

\textbf{``Contour-driven Sumi-e rendering of real photos” \cite{ning2011contour}} paint complex painterly brushstrokes in the
Japanese Sumi-e ink painting style 
through a more complex stroke model. The shapes of the strokes are calculated by spatially clustering the streamlines of an orientation map and replacing the clusters with painted strokes. %

\textbf{``Painterly rendering with content-dependent natural paint strokes” \cite{huang2011painterly}} place multicolored curved brushstrokes according to an importance map, calculated from edge data. The layered painting process is similar to \cite{hertzmann1998} but uses interpolated gradients \cite{litwinowicz}.

\smallskip

\textbf{High-level edges}

\smallskip

These algorithms are not only based on simple pixel edges, but incorporate semantically more meaningful features, such as the contours of actual objects within the image (Figure \ref{fig:image_parsing}).

\textbf{``Painterly rendering using image salience” \cite{collomosse2002}} make use of automatically calculated image salience to place strokes. Salience is calculated through a pre-trained feature detector, which identifies highly visible and statistically surprising image artifacts as important. This salience measure is similar to edge detection but will pick up on more subtle patterns. Brushstrokes are shaped and sized such that important regions are drawn with small, pointy strokes and unimportant regions are painted with large round ones.

\textbf{``Abstracted painterly renderings using eye-tracking data” \cite{santella2002}} add an importance map based on eye tracking data to the algorithm of \cite{hertzmann1998}, which enables the painting to be more detailed in salient regions. They generate a saliency map from a few seconds of eye-tracking data of the target image and use it to enhance the color, contrast, and size of the strokes.

\textbf{``From image parsing to painterly rendering.” \cite{zeng2009image}} parse images through a user-supplied hierarchical segmentation map to facilitate a semantic and object-oriented painting process (Figure \ref{fig:image_parsing}). Semantic segmentation contains not just object boundaries but information about texture and object class as well. Strokes are chosen based on these semantic features from a database of real brushstrokes, such that their texture matches the object to be painted. Their shape is curved along an orientation map, which is calculated through the semantic object contours and low-level edges. The color of the strokes is chosen based on the target image, but can be enhanced by automatically matching the color palette of another painting with similar semantic contents. This painting approach is later adapted to enable more abstract-looking paintings, by systematically perturbing stroke parameters \cite{zhao2010sisley, zhao2013abstract}. 

\textbf{``Video painting with space-time-varying style parameters” \cite{kagaya2010video}} use semantic segmentation to allow varying painting styles between different objects. The strokes are curved along a special tensorfield-based orientation map \cite{zhang2006interactive} from edges and object contours. Painting styles can be chosen individually for each object. These influence the stroke color, position, and orientation. For example, background strokes can be drawn as parallel lines, whereas foreground strokes are more vibrant and follow the semantic edges.

\textbf{``Artistic composition for painterly rendering” \cite{lindemeier2016artistic}} approach painting similarly to \cite{zeng2009image} by segmenting the image first into a hierarchy of foreground and background objects and then painting them individually. For each region, the layered painting process is then very similar to other algorithms \cite{hertzmann1998, lu2010interactive} and aligns the stroke curvature with an interpolated orientation map.

\subsubsection{Region  Approximation}

\label{region_approximation}

\begin{figure}[!t]
\centering
\includegraphics[width=3.5in]{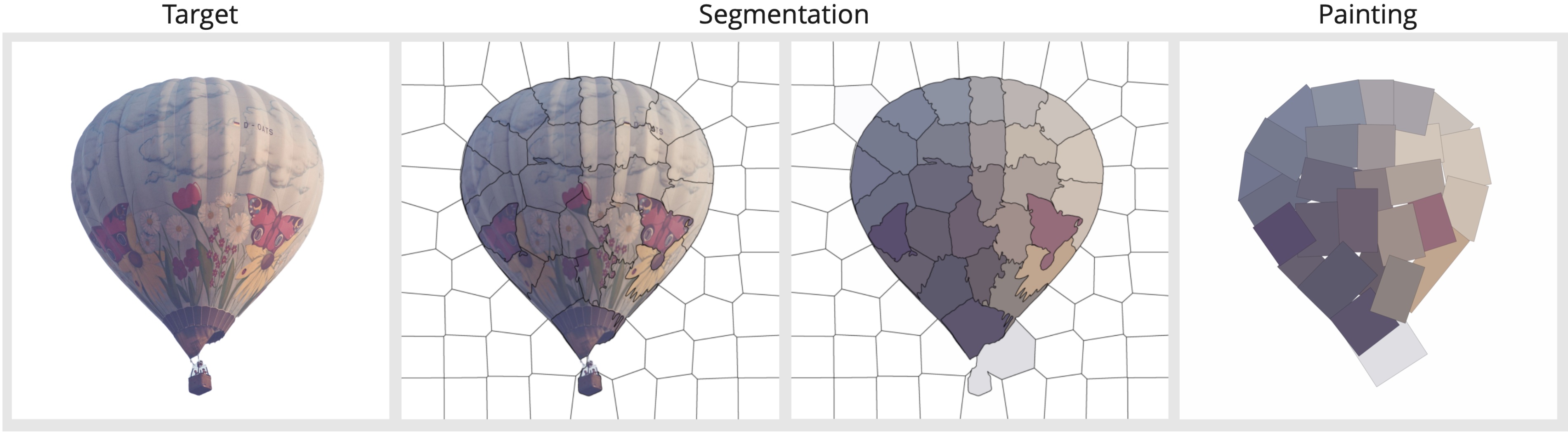}
\caption{Illustration of a region approximation painting algorithm (similar to \cite{song2013abstract}). First, a target image is segmented into regions of similar color. Then, the regions are replaced by matching strokes and colored according to the target image.}
\label{fig:segmentation}
\end{figure}

There are two kinds of region-based SBR approaches: those based on image segmentation and those based on image moments. Both try to find large groups of similarly colored pixels in order to approximate them with brushstrokes of that color (Figure \ref{fig:segmentation}).

\smallskip

\textbf{Image Moments}

\smallskip

Image moments $M$ are a special case of the general concept of a moment from statistics. They are defined as $M_{lm} = \sum_x \sum_y x^ly^mI(x,y)$, where I is a grayscale image \cite{shiraishi2000algorithm}. Image moments can be used to calculate useful properties of objects in images, most notably their center of mass, size, and orientation. In this context, an object is defined as having dark pixels on a light background. A region in an image can be approximated through strokes by replacing it with one or more brushstrokes of the same image moments as the target.

\textbf{``An algorithm for automatic painterly rendering based on local source
image approximation” \cite{shiraishi2000algorithm}} are the first to present an automatic SBR algorithm solely based on image moments. They specifically try to incorporate more color information instead of relying on gradients and segments, which are often determined using grayscale images. To accomplish this, they calculate the image moment of a region with respect to a given color. This enables them to place a colored stroke with the equivalent image moments on the canvas. Importantly, they are not searching for connected regions of similar color within the image, but instead use the overall color distribution of the image to guide stroke placement. Their approach results in the placement of a small number of large strokes in regions of similar color and many small strokes in areas with multiple different colors.

\textbf{``Multiscale moment-based painterly rendering” \cite{nehab2002multiscale}} extend \cite{shiraishi2000algorithm} in two ways. First, they give more control over the distribution of stroke positions. This enables their second addition, which is a multi-layered painting strategy. They start by drawing large strokes in the background of the image and add smaller strokes only in regions with strong edges and small image moments.

\smallskip

\textbf{Image Segmentation}

\smallskip

Image segmentation is the task of grouping neighboring pixels in an image, similar to clustering in statistics or machine learning \cite{szeliski2010computer}. The final segmentation maps usually consist of a number of compact, disjoint regions without holes. Figure \ref{fig:segmentation} shows a segmented photograph with its stroke-based approximation. There are countless approaches to automatic image segmentation, with many being based on the idea of recursively splitting or merging regions \cite{szeliski2010computer}. The main idea behind approximation through segmentation is to replace an image region with a similar-looking brushstroke. The main task is to find a stroke of similar size, shape and orientation as the image segment. The stroke color can simply be taken as the mean or median pixel color of the region. Segmentation algorithms usually have some parameter that governs the size and number of segments, which can be used to paint large strokes in the background and small ones in the foreground \cite{song2013abstract, gooch2002artistic}. Regions of high importance can be segmented into more regions in order to retain detail \cite{gooch2002artistic}. It should be noted that there are several rendering approaches that use colored segmentation maps as a painting \cite{wen2006color, decarlo2002stylization}. Since these methods do not calculate simple stroke parameters, they are not considered in this survey.

\textbf{``Artistic vision: painterly rendering using computer vision techniques” \cite{gooch2002artistic}} use pixel-based computer vision algorithms to calculate the medial axis of small image segments. This medial axis represents the center line of the region and can be approximated through a curved spline brush. User-selected regions can be segmented with a higher level of detail, resulting in a larger number of small strokes.

\textbf{``Empathic painting: interactive stylization through observed emotional state” \cite{shugrina2006empathic}} use not one, but many curved strokes to paint large image segments. This is done by first filling in their interior with multiple parallel strokes and then tracing the contour with long, curved lines. The stroke width is proportional to the segment size, and all control points of the curves are randomly jittered to reduce the regularities in the painting. Additionally, they provide methods to change stylistic parameters such as color, jitter, and curvature based on the emotions observed in a live video stream.

\textbf{``Emotionally aware automated portrait painting” \cite{colton2008emotionally}} use a similar approach to \cite{shugrina2006empathic}, with large regions containing multiple strokes and stylistic parameters being changed according to high-level input. They provide additional filling methods for the image segments and enable layered painting through repeated segmentation at different scales, stacked on top of each other.

\textbf{``Abstract art by shape classification” \cite{song2013abstract}} use a small number of large shapes instead of many curved lines to match the image segments. Through hierarchical image segmentation, they divide the picture into large, compact shapes and match them with primitives such as circles, triangles, rectangles, superellipses, and convex hulls. The best shape parameters are chosen through various techniques, such as image moments, depending on the primitives. They paint multiple layers based on differently sized segments by starting with big shapes and only drawing smaller strokes on top if their color significantly deviates from the background. Different styles can be realized through the combination of different primitives. These individual stylistic decisions can even be learned through a shape classifier.

\textbf{``A generic framework for the structured abstraction of images” \cite{faraj2017generic}} depend on the special properties of the so-called ``tree of shapes" \cite{monasse2000fast}
for their painting approach. 
Here, hierarchical overlapping image segments are calculated on the basis of the luminance of the image. The tree of shapes is able to correctly handle regions with more complex topology, where regular segmentation algorithms would fail. A painterly image is rendered by replacing the shapes of the tree with fitting primitives. Additional effects can be achieved by blurring, smoothing, and displacing the shapes.

\subsection{Painting through error minimization}
\label{error_minimization}

Error minimization techniques do not specify exactly how the stroke parameters should be calculated, but how the final image should look from a top-down perspective \cite{hertzmann2001paint}. This is achieved by viewing the painting procedure as the minimization of an error function with respect to the stroke parameters through iterative optimization or deep learning. The error or loss function should in some way inform about the achieved quality of the painting. There are many different losses available, but almost all algorithms try to minimize some kind of pixel difference between the rendered painting and a target image. The preferred painting style can be controlled
through the loss function \cite{kang2006unified} (Figures \ref{fig:balloon_white_loss}, \ref{fig:balloon_loss}, \ref{fig:mona_lisa_loss} and \ref{fig:style_transfer}) and the stroke model (Figure \ref{fig:brushes_paintings}). Even interactions between strokes can be easily managed \cite{zhao2011customizing}.

There are two types of error minimization algorithms. Optimization algorithms (section \ref{iterative_optim}) iteratively change stroke parameters until a low error is achieved. This can be accomplished through the use of randomized search, genetic algorithms or gradient descent. Machine learning algorithms (section \ref{DL_painting}) do not only optimize parameters for one painting but try to learn how to paint arbitrary images. This is done by training neural networks to predict the parameters of brushstrokes through supervised or reinforcement learning.

\subsubsection{Objectives for optimization}
\label{losses}
Despite their algorithmic differences, most learning and optimization approaches can use similar metrics to evaluate the quality of a given painting $I_P$. These can not only enforce a good stroke-based reconstruction of the target image $I_T$ but also encode stylistic concepts (Figure \ref{fig:style_transfer}). Additionally, they can enable the models to achieve other tasks such as text visualization or unconditional image generation (Figure \ref{fig:spiral_unconditional}). Depending on the algorithm, the objective might be called ``loss", ``error", or ``energy" and in the context of genetic algorithms, they are inverted into so-called ``fitness" scores. For simplicity, we will refer to these objectives as loss functions $L$. 
Compound target functions can be created through a weighted sum of multiple different losses. For example, $L = \lambda_1 L_{style} + \lambda_2 L_2$ \cite{zou2021stylized}.

In colored images, representations of pixel values differ, depending on the color space that is used \cite{szeliski2010computer}. Common examples are RGB, XYZ, CIELAB, and HSV color spaces, which all use three values to define the color of a pixel. The color space can have an impact on the performance of models \cite{gowda2018colornet}, but in practice, most authors simply use RGB images. 

In the following, we describe the most common losses for painterly rendering algorithms. The exact formulas and details may differ between implementations.

\begin{figure}[!t]
\centering
\includegraphics[width=3.5in]{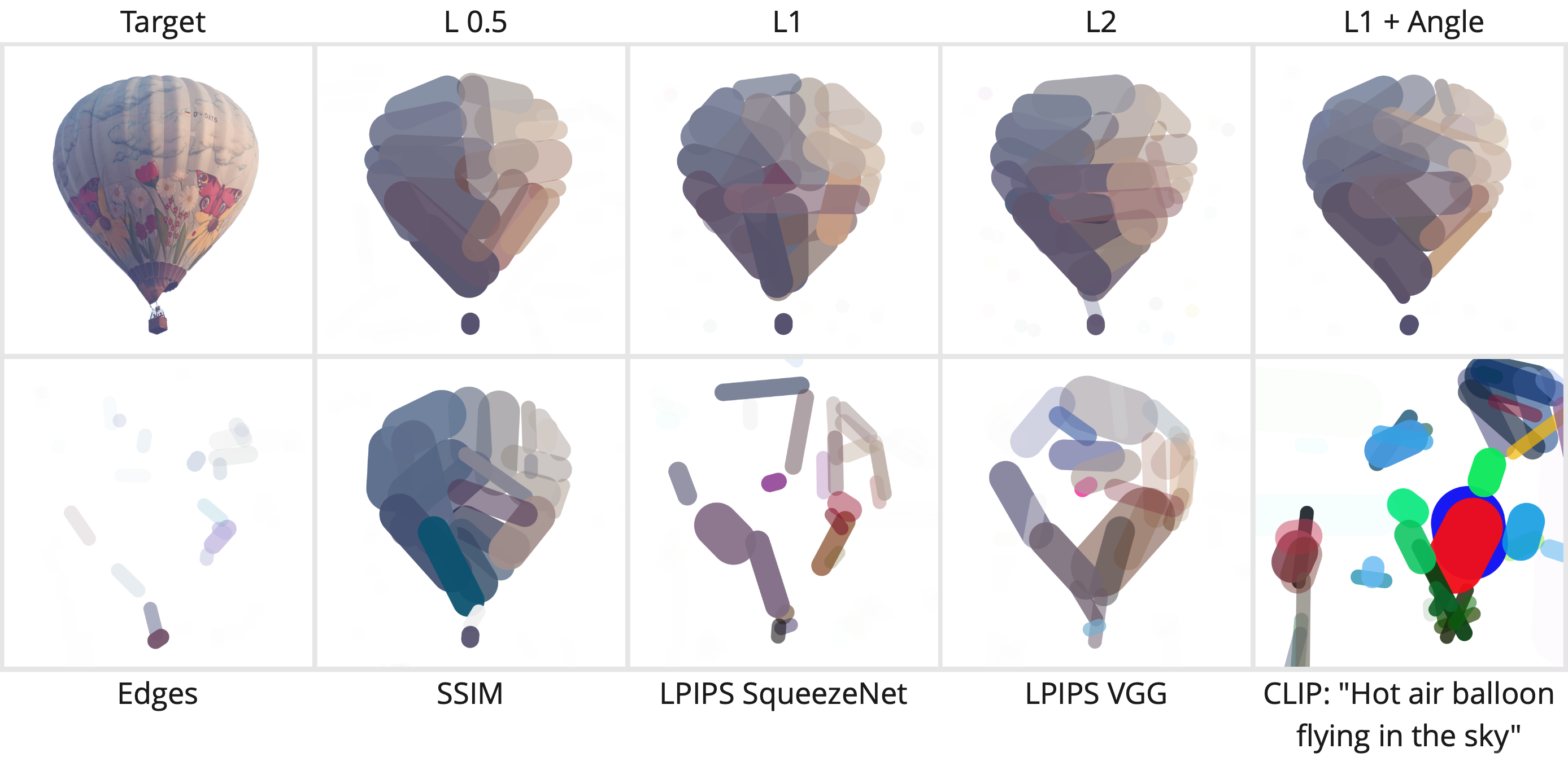}
\caption{Optimization of 50 lines with different loss functions. Due to the very limited number of strokes, not all loss functions lead to a good painted result.}
\label{fig:balloon_white_loss}
\end{figure}

\begin{figure}[!t]
\centering
\includegraphics[width=3.5in]{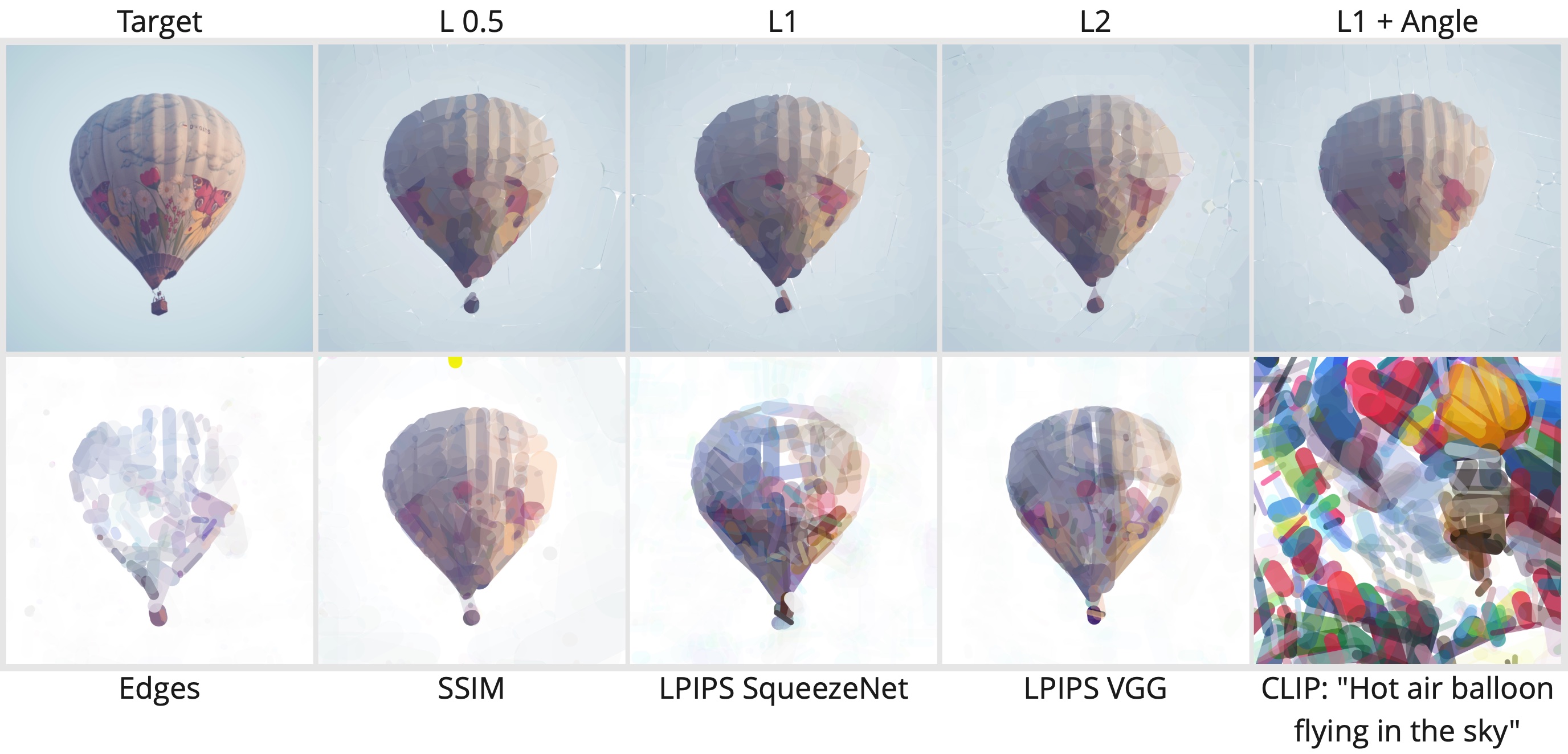}
\caption{Layered painting of 500 lines in two layers, with different loss functions. All $L_p$ losses are able to generate a good reconstruction of the target image.}
\label{fig:balloon_loss}
\end{figure}

\begin{figure}[!t]
\centering
\includegraphics[width=3.5in]{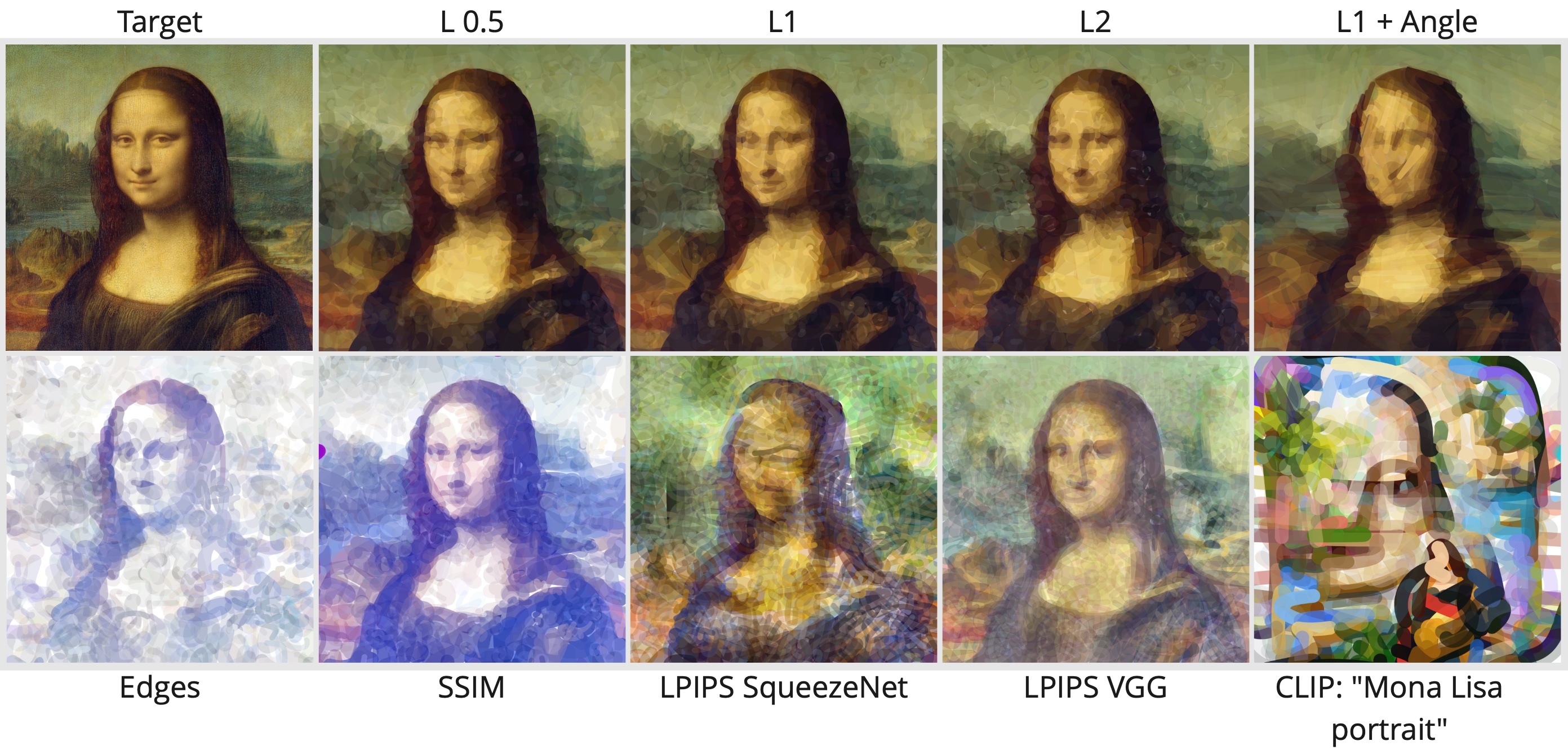}
\caption{Layered optimization of 1000 curves in three layers, with different loss functions. While the perceptual losses do not produce paintings that closely follows the target colors, the overall image structure is retained.}
\label{fig:mona_lisa_loss}
\end{figure}

In addition, we provide overviews  of the effects of different loss functions on the appearance of paintings. For this, we extend the stroke optimization by \cite{li2020differentiable} and \cite{frans2021clipdraw} into a layered painting algorithm (Figure \ref{fig:layers}) that can optimize for a variety of loss functions (figures \ref{fig:balloon_white_loss}, \ref{fig:balloon_loss}, \ref{fig:mona_lisa_loss}, \ref{fig:style_transfer}).

\textbf{Stroke losses} can be used to guide the style of a painting by directly influencing the values of the stroke parameters, regardless of the target image. There are many different types of possible stroke losses. For example, the loss $L(S)$ might be proportional to the number, size, or curvature of strokes \cite{hertzmann2001paint, collomosse2005genetic}. The interaction of strokes can also be incorporated by measuring how much the strokes overlap \cite{kang2006unified, wang2021objects} and how similar the parameter values of neighboring strokes are \cite{kotovenko2021rethinking, zhao2011customizing}. In the figures \ref{fig:balloon_white_loss}, \ref{fig:balloon_loss}, \ref{fig:mona_lisa_loss}, we use the angle between neighboring strokes as a loss, in order to achieve 
a bias towards aligned stroke orientations.

\textbf{\boldmath$L_p$-losses} are a natural choice to measure the distance between two images $I_P,I_T$ of size $M\cdot N$ by simply comparing their individual pixel values:
$$L_p(I_P, I_T) = \sum_{px} ||I_P - I_T||_p$$

With $p = 2$, this is equivalent to optimizing for the squared error between images. Other values, such as $p = 1$ or $p = \frac{1}{2}$ \cite{jia2019paintbot} can be used as well and might sometimes result in paintings with slightly better-defined shape boundaries, but slightly worse average color matching \cite{jia2019paintbot}. In the following, we loosely use the terms $L_2$ and $L_1$ to refer to squared and absolute errors respectively. In order to reduce the impact of textures and noise, images and paintings are sometimes blurred before calculating the loss \cite{mihai2021differentiable, kang2006unified}. Although they are used at least as a baseline in almost all algorithms, $L_p$ losses are often cited as having an unfavorable flat loss landscape and being outperformed by more sophisticated perceptual and adversarial losses \cite{huang2019learning, zou2021stylized, mihai2021differentiable, zhang2018unreasonable}.

\textbf{Perceptual losses} \cite{zhang2018unreasonable} use neural networks, trained for image classification tasks \cite{simonyan2014very, iandola2016squeezenet}, to measure the similarity between images. The basic idea is that images with similar content should result in a similar layer activation. The similarity of images $I_1$ and $I_2$ can, for example, be calculated from the feature maps $\phi$ of layer $j$ with shape $C \times H \times W$ as:

$$L_{Perc}^j(I_P, I_T) =  \frac{1}{CHW}||\phi_j(I_P) - \phi_j(I_T)||^2_2$$

Early layers $j$ will retain colors, textures, and shapes, while higher layers  preserve the high-level content and image structure \cite{johnson2016perceptual}. \cite{nolte2022stroke} show that perceptual losses can benefit from being trained on stroke-based images, which can lead to more useful and less flat gradients.

\textbf{Style losses} are used within the context of neural style transfer \cite{gatys2016image, johnson2016perceptual} (Figure \ref{fig:style_transfer}) not to measure content, but the visual look, such as the painting style of images. Without going into too much detail, the idea is that the style of an image is encoded in the distribution and correlation of layer activations. In the original implementations, this is measured by the Gram matrix $G_j$ \cite{gatys2016image, johnson2016perceptual} and the style loss can then be calculated as:
$$L_{Style}(I_P, I_T)=||G_j^{\phi}(I_P)-G_j^{\phi}(I_T)||^2_F$$
However, this is not the only way to calculate a neural style loss \cite{jing2019neural}. Regardless of implementation, it is usually used together with other perceptual losses that encode the content of the image\cite{schaldenbrand2022styleclipdraw, zou2021stylized, kotovenko2021rethinking}: $L = \lambda_{1}L_{Content}(I_P, I_{T_1}) + \lambda_{2}L_{Style}(I_P, I_{T_2})$.

\textbf{Adversarial losses} \cite{goodfellow2014generative, arjovsky2017wasserstein} arise from the "adversarial" competition between two neural networks: a generator network $G$ aims to create an image that a second network $D$ ("discriminator") cannot recognize as being synthesized ("faked" by the generator). Adversarial loss measures the expected advantage of one network over the other in a game when the discriminator is randomly fed with "true" images or "fake" images from the generator while both networks adapt towards
their opposite goals of "faking" and "discriminating". 

Training using Adversarial losses allows to obtain generator networks that generate unconditional images, without an explicit target image (Figure \ref{fig:spiral_unconditional}). Different adversarial losses are possible, leading to different similiarity metrics in the
space of generated images. A particular stable loss is the  Wasserstein (WGAN) loss \cite{arjovsky2017wasserstein} that leads to a well-behaved distance between the real and artificial images, resulting in more stable gradients.\\

$L_{Adv} = D(I_P)$ or $L_{Adv} = D(I_P | I_T)$\\

\textbf{CLIP} (Contrastive Language–Image Pre-training)\cite{radford2021clip} \cite{jia2021scaling} losses allow optimization with target text prompts instead of target images. Two encoders $f$ and $g$ take an image $I$ and text $T$ to generate text and image embeddings: $f(I) \in R^d, g(T) \in R^d$. The encoders are trained on a large set of  image and text pairs and judged through the similarity of their embeddings. Real text-image pairs should have very similar embeddings, whereas false pairs should have dissimilar ones.  The loss of a text-image pair can be calculated e.g. through a cosine distance of the output of the final trained encoders:
$$L_{CLIP} = 1-\frac{f(I)^T g(T)}{||f(I)||_2 \, ||g(T)||_2}$$
When using a CLIP loss, it is often beneficial to take the average loss over multiple transformed and augmented versions of the same painting, instead of computing the loss only a single time \cite{frans2021clipdraw, tian2022modern}.

\smallskip

Apart from these losses, all sorts of image metrics such as the structural similarity \cite{wang2004image} or the $L_1$ difference of edge intensities can be used to guide a stroke optimization (figures \ref{fig:balloon_white_loss}, \ref{fig:balloon_loss}, \ref{fig:mona_lisa_loss}).

\subsubsection{Iterative parameter optimization}
\label{iterative_optim}

\begin{figure}[!t]
\centering
\includegraphics[width=3.5in]{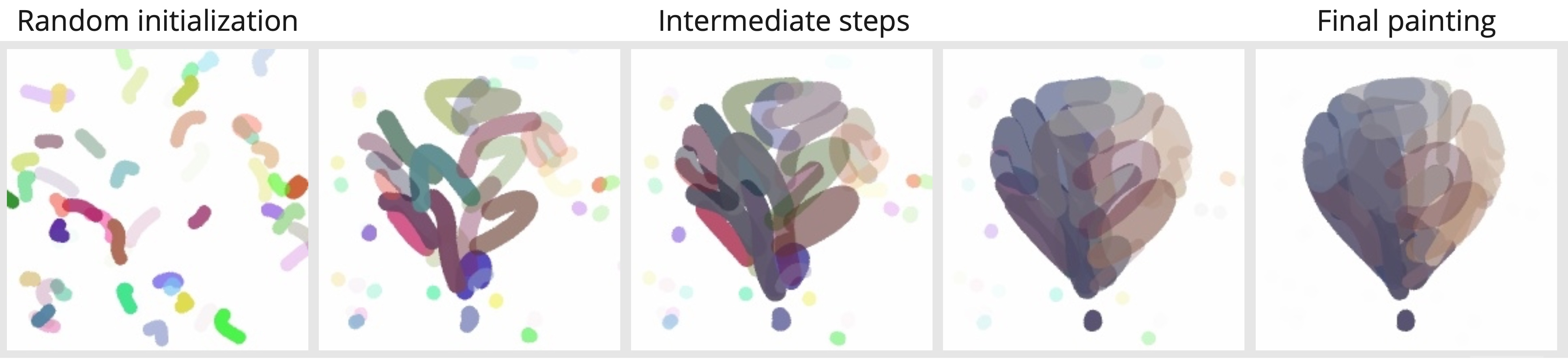}
\caption{Some of the individual steps when optimizing the parameters of 50 curved strokes through gradient descent based on the algorithms of \cite{li2020differentiable} and \cite{frans2021clipdraw} with an $L_1$ loss.}
\label{fig:optim_steps}
\end{figure}

\begin{figure}[!t]
\centering
\includegraphics[width=2.5in]{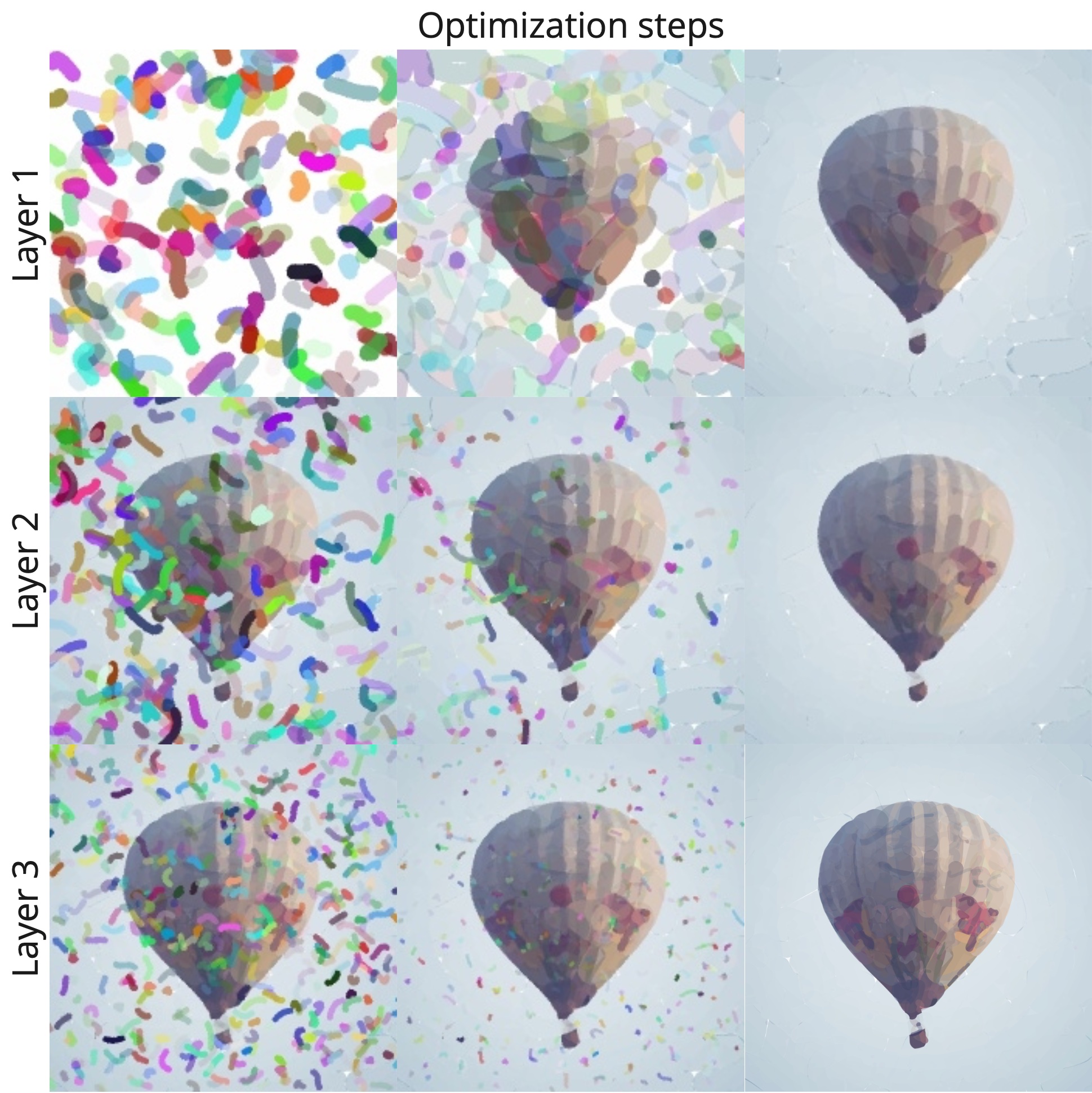}
\caption{Layered extension of the stroke optimization by \cite{li2020differentiable, frans2021clipdraw} with 1000 strokes and an $L_1$ loss, inspired by the coarse-to-fine painting procedure of human painters. After one layer of strokes is painted, the parameters are frozen and the next layer of strokes is randomly initialized and optimized on top of the underlying layer.}
\label{fig:layers}
\end{figure}

\begin{figure}[!t]
\centering
\includegraphics[width=3.5in]{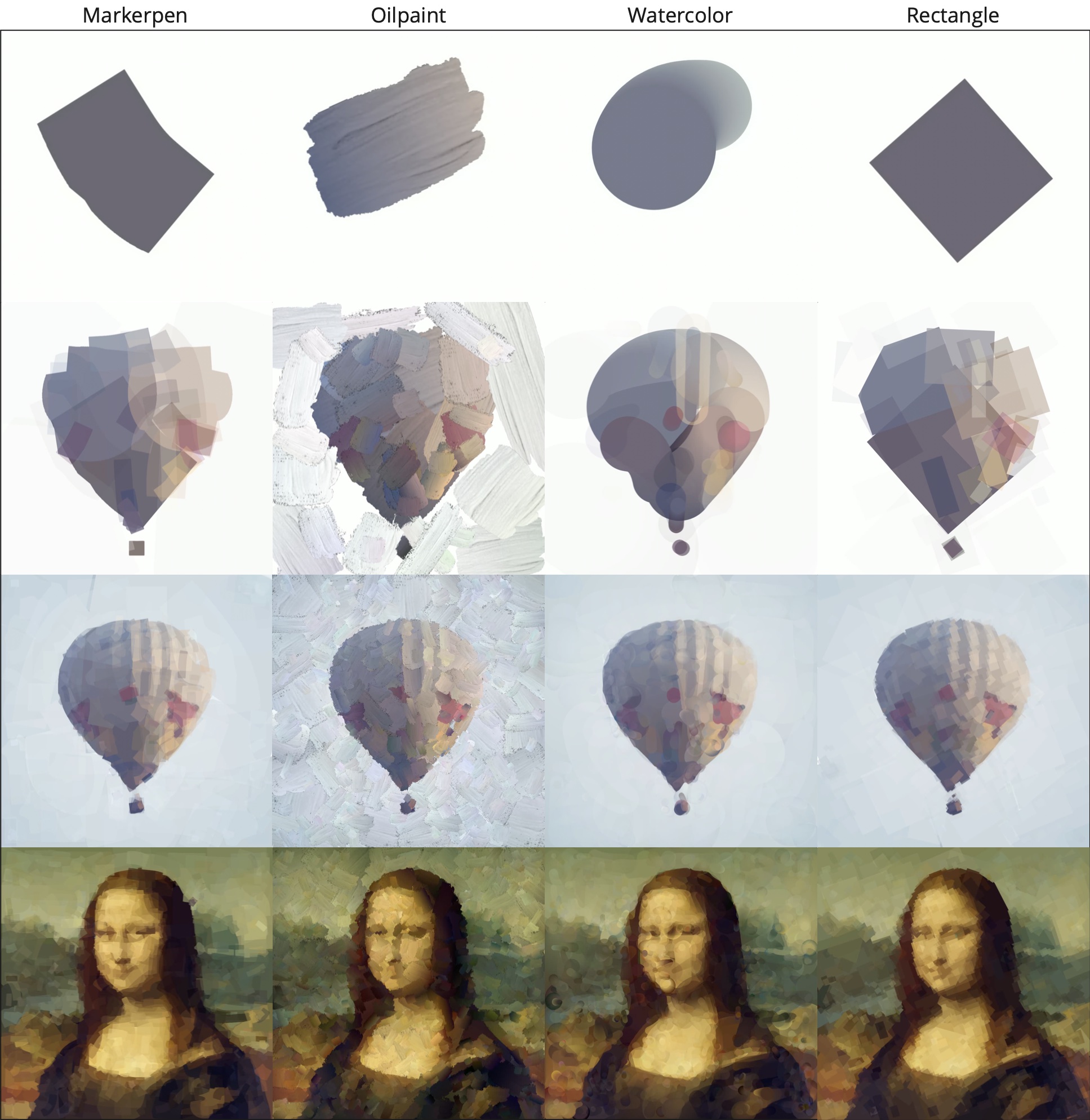}
\caption{Comparison of the different stroke models from \cite{zou2021stylized}, painted with 50, 500 and 1000 strokes (top to bottom). Different painting styles can be achieved by varying the stroke model and number of brushstrokes.}
\label{fig:brushes_paintings}
\end{figure}

\begin{figure}[!t]
\centering
\includegraphics[width=3.5in]{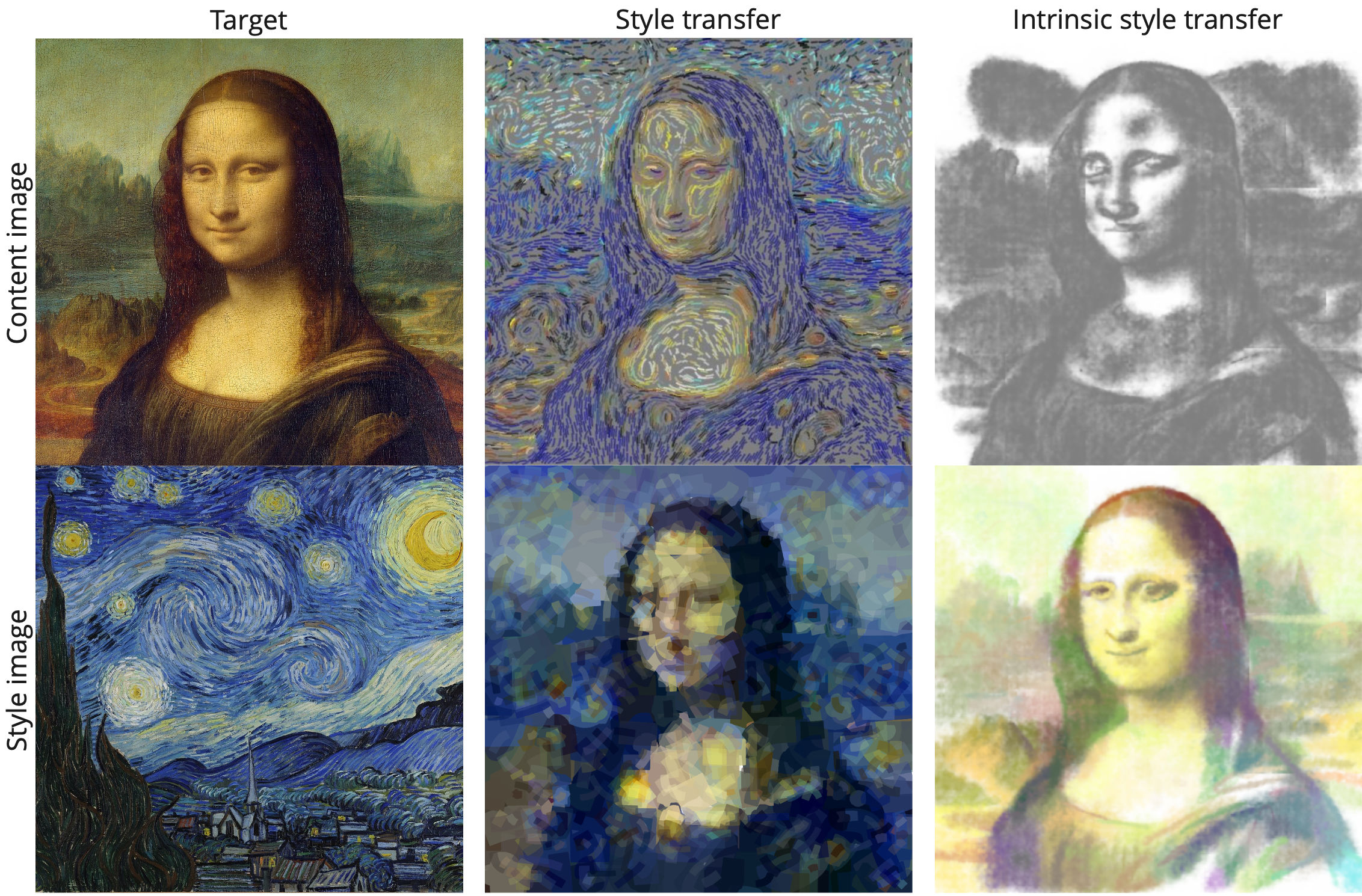}
\caption{Results of style transfer with different optimization algorithms. Stroke parameters are optimized through perceptual VGG losses to imitate the style of a target painting (left) while retaining its content. Middle-top: Kotovenko et al. \cite{kotovenko2021rethinking}, middle-bottom: Zou et al.  \cite{zou2021stylized}, using the ``Markerpen" stroke model. Right: ``Intrinsic style transfer" from \cite{nakano2019neural}, where only the content loss is optimized and the style is dictated by the stroke model, with watercolor and grayscale brushes.}
\label{fig:style_transfer}
\end{figure}

\begin{figure}[!t]
\centering
\includegraphics[width=3.5in]{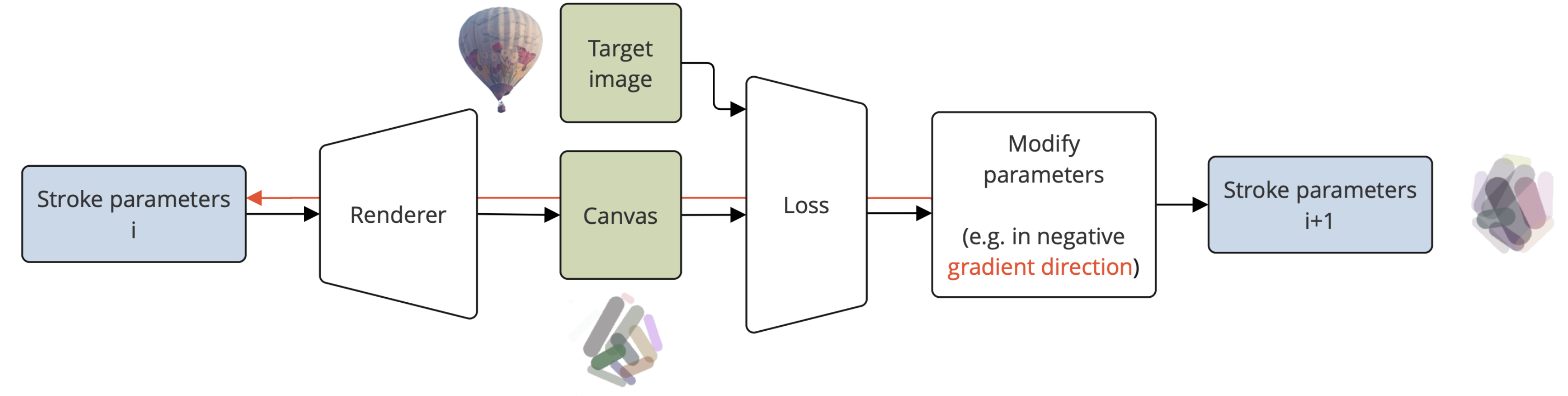}
\caption{Optimization algorithms iteratively modify stroke parameters until the canvas looks like the target image. Gradient descent needs to be able to calculate the gradient of the loss value with respect to stroke parameters (red) to propose updates for the stroke parameters. This is only possible if the loss and renderer are differentiable. Randomized search algorithms omit the gradient calculation and instead modify the stroke parameters through stochastic techniques such as mutation and genetic selection.}
\label{fig:gradient_descent}
\end{figure}

We divide iterative painting methods into three distinct categories: brute-force search, randomized search, and gradient descent. The basic idea behind all these algorithms is to iteratively search for stroke parameters that produce a painting with a low loss or error value (Figure \ref{fig:optim_steps}). Due to their iterative nature and often costly computations, these methods are usually slower than greedy or learning-based painting algorithms. Brute-force search can only be used on very limited, discrete parameter spaces, but gradient descent and randomized search can be used to find continuous values for larger parameter spaces.
Gradient descent requires the computation of derivatives of the loss with respect to painting actions (Figure \ref{fig:gradient_descent}). This is usually only possible through differentiable rendering (section \ref{differentiable_rendering}).%

\smallskip

\textbf{Brute-force search}

\smallskip

Brute-force painting algorithms exhaustively search for optimal stroke parameters over a limited number of discrete values. Searching over a whole parameter space is only computationally feasible if it is reasonably sized. Due to this, brute force approaches are usually constrained to only optimizing individual parameters and strokes, instead of combinations of stroke parameters. For example, they may search for the best next stroke to put down in a small region on the canvas. Or they search for the optimal rotation of a stroke, given some color and position. For this, they calculate all possible parameter values and select those that result in the lowest image loss. Those parameter values that are not optimized are often instead calculated through the use of greedy strategies, such as described in the section \ref{greedy}.

\textbf{``Paint by relaxation” \cite{hertzmann2001paint}} optimizes curved brush strokes by using relaxation. Strokes are first painted according to a greedy strategy \cite{hertzmann1998} and then optimized like active contours, or "snakes"\cite{kass1988snakes}. Then they are judged through the use of a weighted energy function which encourages them among other things to be smooth, follow the image edges, and minimize the $L_1$ loss. Optimization is done by repeatedly searching for the best control point positions in a limited pixel neighborhood around the strokes.

\textbf{``Anipaint: Interactive painterly animation from video” \cite{Donovan2011anipaint}} introduce a fast and efficient stroke optimization procedure by calculating a compound energy function for individual strokes, without regard for stroke interaction. The stroke energy is minimized if a curved stroke smoothly follows a segmentation-based orientation field and matches the color of the target image. The strokes are grown from a starting location, with one control point after the other being placed to achieve the lowest possible energy value.

\textbf{Stroke database algorithms \cite{seo2013interactive, seo2009painterly, lee2021stroke}} paint through the use of stroke-databases that contain a large number of photo-scanned strokes with different shapes, textures, and rotations. Strokes are semi-randomly distributed on the canvas in layers and colored according to the target image. For every stroke position, that stroke from the database, which results in the smallest $L_2$ error to the target image, is chosen. Different painting styles can be achieved by modifying or limiting the database and including a chance for the algorithm to place sub-optimal strokes \cite{seo2013interactive}.

\textbf{``Portrait painting using active templates” \cite{zhao2011portrait}} take painting with a database a step further by building a collection of portraits, painted by artists with parameterized strokes. Given a portrait photo, they select the best-fitting painted template from the database according to its shape and color. It is then further transformed to match the proportions and appearance of the target portrait more closely.

\smallskip

\textbf{Randomized search}

\smallskip

The first and perhaps simplest stochastic optimization algorithm for painterly rendering was proposed in the seminal work of Haeberli \cite{haeberli}. Starting from an arbitrary initialization, the parameters of multiple brushstrokes are incrementally perturbed by small amounts. Changes are only accepted if they result in a solution with a decreased $L_2$ distance to the target image. This is an example of a hill-climbing algorithm. Other popular options for realizing randomized searching include simulated annealing and evolutionary algorithms.

Simulated annealing works the same as hill climbing, but changes to the parameters that \textit{raise} the loss are accepted with a small probability as well. This probability is equal to $e^\frac{-\Delta Loss}{T}$, where $T$ is a temperature parameter which decreases over the iterations \cite{kirkpatrick1983optimization, berg2019evolved, kang2006unified}. In the beginning, a high temperature leads to a search over a larger parameter space and makes the method less prone to getting stuck in a local optimum. 

Evolutionary and genetic algorithms have a long history of being used in computer-aided image generation \cite{sims1991artificial, latham1992evolutionary} and are a popular choice for optimizing SBR parameters due to their ability to handle large and complex parameter spaces \cite{collomosse2005genetic}. They optimize the parameters of arbitrary functions by mimicking biological evolution. In SBR, a painting is made up of a sequence of stroke parameters that can be interpreted as a genome. Genetic algorithms mutate and combine the stroke parameters of many individual paintings with the goal of increasing their fitness score. This fitness score is some measure of similarity between the target image and the rendered painting of an individual. Genetic algorithms gradually converge to having many individuals with a high fitness score by slightly randomly mutating stroke parameters, discarding the worst performers of each generation, and carefully combining the parameters of the best paintings into new individuals. The different genetic algorithms for SBR mainly differ in their genome representation and how they implement mutation and reproduction.

In general, many different randomized search algorithms can be used to find good stroke parameters, and there is not one clear best choice. For example, \cite{chakraborty2007image} show that their genetic algorithm finds better optima than simple hill climbing, while \cite{paauw2019paintings} conclude that hill climbing will produce better results in the same number of iterations and that the performance of simulated annealing strongly depends on the choice of the temperature $T$.

\textbf{Random changes to a painting:}

All of these algorithms optimize their objective by randomly changing a painting. Their performance and results depend, among other things, on how exactly they implement these mutations. Within the context of SBR, this does not mean direct pixel changes but changes to the stroke parameters, which in turn result in changes to the rendered image. It is sometimes useful or necessary to control how large a random change in an image is. The impact of a modification can be restricted by limiting the number of parameters that are changed simultaneously and by limiting the maximum amount of change for the values of individual parameters \cite{paauw2019paintings}.

Which alterations are possible depends on the stroke model (section \ref{stroke_definition}). In the following, we describe several possibilities for randomly modifying stroke-based images.

\smallskip

\textbf{Parameter modifications:} 
\begin{itemize}
    \item Mutate one or more stroke parameters. Depending on the stroke model, this can change e.g. the size, shape, orientation, color, texture or position of a stroke.
    \item Add a new stroke, randomly or through a greedy algorithm
    \item Delete a stroke
    \item Add or remove control points from a curved stroke or polygon
    \item Change the painting order of some strokes
\end{itemize}

Many algorithms intentionally leave some parameters out of the optimization and do not mutate them to simplify the parameter search. For example, stroke orientation is sometimes not optimized, but is chosen to align with the edges of the image \cite{collomosse2005genetic, chakraborty2007image, kang2006unified}.

\smallskip

\begin{figure}[!t]
\centering
\includegraphics[width=3.5in]{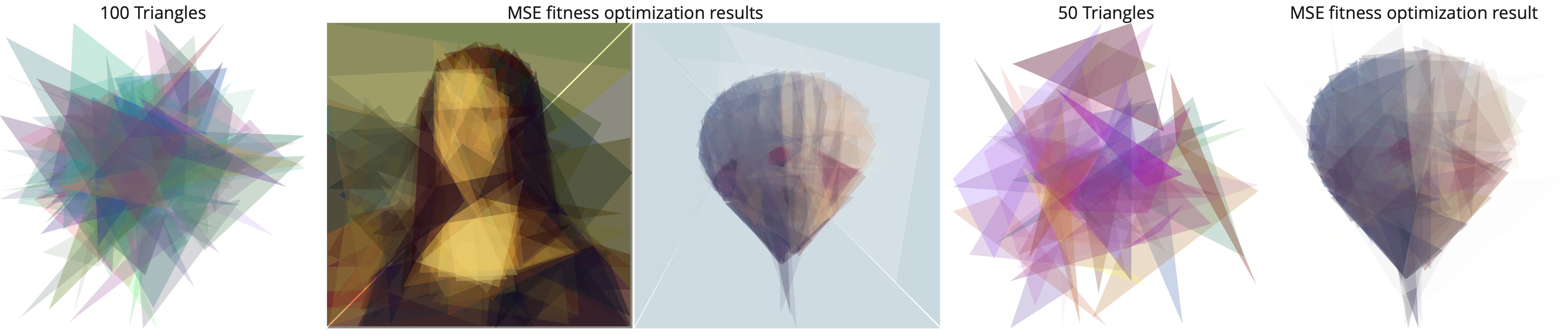}
\caption{Results of genetic polygon optimization, generated with the implementation of \cite{tian2022modern}. Initialization and optimization results with 100 triangles (left) and 50 triangles (right).}
\label{fig:polygon_optimization}
\end{figure}

\textbf{``Random paintbrush transformation” \cite{sziranyi2000random}} implement hill climbing by repeatedly generating random strokes and only painting them on the canvas if they decrease the $L_2$ loss to the target. Unlike other hill climbing algorithms \cite{haeberli, paauw2019paintings}, parameters are not modified after they have been selected. They later optimize this procedure by using MCMC sampling, so that the stroke parameter proposals are chosen based on the current canvas, resulting in faster convergence and fewer overlapping strokes \cite{sziranyi2001optimization}.

\textbf{``Genetic paint: A search for salient paintings” \cite{collomosse2005genetic}} choose to optimize for saliency and detail preservation through a genetic algorithm. Individual stroke parameters are decided according to greedy edge alignment (section \ref{edge_alignment}) with some stochastic variation. The evolutionary process then selects and combines those strokes for the next generation, which deliver the closest matching saliency and detail compared to the target image. 

\textbf{``A unified scheme for adaptive stroke-based rendering” \cite{kang2006unified}} present a unified approach to achieve a broad class of illustration styles by optimizing stroke parameters with simulated annealing. This is done by mapping painting styles to various hyperparameter choices that influence the optimization process. For example, mosaic and pointillist styles can be achieved by penalizing overlapping strokes through the loss function. Different importance and orientation maps can be used to vary the size and orientation of the strokes.

\textbf{``Image-based painterly rendering by evolutionary algorithm” \cite{chakraborty2007image}} use a multilayered genetic stroke optimization where not just stroke parameters, but also the number and order of strokes in each layer are explicitly optimized. This enables their algorithm to find better optima than simpler hill-climbing algorithms.

\textbf{Genetic polygon optimization algorithms\cite{bergen2012automatic, berg2019evolved, paauw2019paintings}} paint images by optimizing the parameters of polygons and other strokes to achieve a low distance to a target image. Painterly rendering with polygons is NP-hard \cite{van2020simplified} but enables a distinct aesthetic through its complex shapes (Figure \ref{fig:polygon_optimization}).

\textbf{``Incremental Evolution of Stylized Images” \cite{uhde2021incremental}} present an incremental, layered painting approach, which simplifies the search space for its genetic algorithm. Instead of handling and refining all strokes at once, the painting is constructed through individually optimized layers.%

\textbf{``Generative art using neural visual grammars and dual encoders” \cite{fernando2021generative}} use a genetic algorithm to generate images that maximize the CLIP similarity to a text prompt. Instead of directly optimizing the parameters of strokes, the genetic algorithm mutates and optimizes the parameters of multiple hierarchically arranged LSTM models which convert the input string into a sequence of stroke parameters. 

\textbf{``Modern evolution strategies for creativity: Fitting concrete images and abstract concepts” \cite{tian2022modern}} (Figure \ref{fig:polygon_optimization}) show that the evolutionary strategy PGPE \cite{sehnke2010pgpe} achieves comparable and even superior performance when optimizing polygons through a non-differentiable renderer, compared to gradient-based optimization with differentiable rendering. Their method works with L2 image fitness as well as CLIP text similarities. 

\smallskip

\textbf{Gradient-based optimization}

\smallskip

\begin{figure}[!t]
\centering
\includegraphics[width=2.5in]{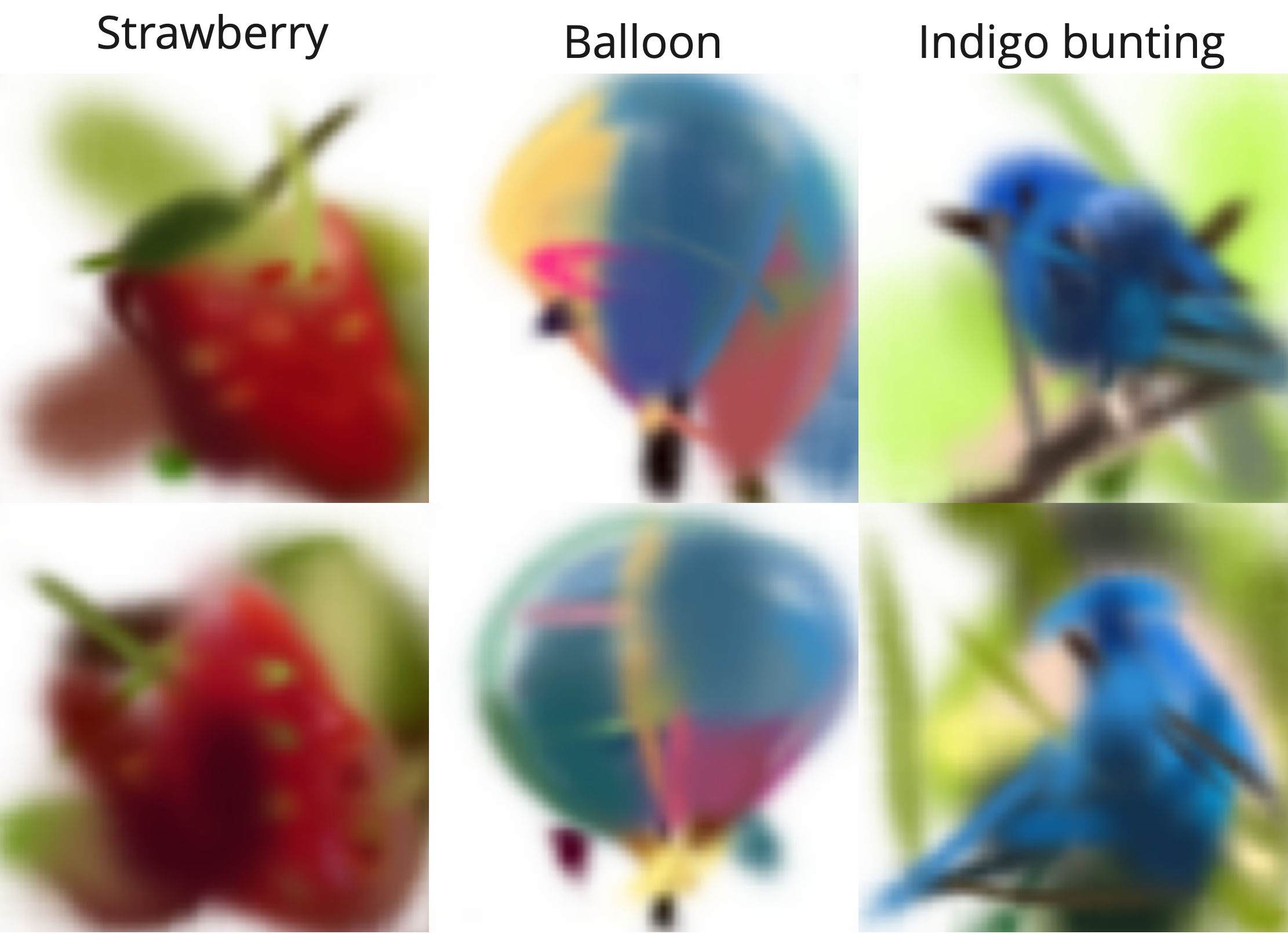}
\caption{Strokes can not just be used to paint target images. \cite{nakano2019neural} optimizes stroke parameters to maximally activate a neuron of a pre-trained image classifier, in order to visualize ImageNet classes \cite{imagenet}.}
\label{fig:imagenet}
\end{figure}

\label{gradient_optim}
Differentiable rendering (section \ref{differentiable_rendering}) enables the use of efficient gradient-based optimization methods, such as stochastic gradient descent (Figure \ref{fig:gradient_descent}), by allowing the computation of the derivative $\frac{\partial L}{\partial s}$ of pixel-based loss functions $L(I_P, I_T)$ with respect to stroke parameters $s$. Gradient descent takes a step in the negative gradient direction of the loss function, scaled by a step size $\gamma$: $S_{n+1} = S_n - \gamma \nabla L(S_n)$. In practice, most algorithms use stochastic gradient descent algorithms such as Adam \cite{kingma2014adam}. 

\textbf{``Customizing painterly rendering styles using stroke processes” \cite{zhao2011customizing}} achieve various painting styles by iterating a reaction-diffusion process on a stroke neighborhood graph. Using differentiable rendering is cleverly avoided in this algorithm, by only calculating derivatives in the parameter space, and never based on pixel values. The proposed method does not calculate gradients from a loss function but uses stochastic reaction diffusion to optimize the interaction and contrast of stroke colors, sizes, and orientations, according to intuitive style parameters.

\textbf{``Neural painters: A learned differentiable constraint for generating brushstroke paintings” \cite{nakano2019neural}} optimize the stroke parameters through a differentiable neural renderer and use pre-trained neural networks to calculate multiple different image losses. Networks trained for image classification are utilized to paint distinct categories of images without a target image (Figure \ref{fig:imagenet}). Alternatively, the loss of content from neural style transfer models \cite{gatys2016image} is used to draw reference images where the style of the painting is not dictated by a target image, but by the stroke model itself (Figure \ref{fig:style_transfer}).

\textbf{``Differentiable vector graphics rasterization for editing and learning” \cite{li2020differentiable}} present a simple painterly rendering algorithm, which works by using their differentiable vector graphics rasterizer. Starting from a completely random initialization, the parameters of a large number of curved Bézier strokes and other shapes can be efficiently adapted through stochastic gradient descent to minimize $L_2$ and perceptual losses. Figures \ref{fig:balloon_white_loss}, \ref{fig:balloon_loss}, \ref{fig:mona_lisa_loss}, \ref{fig:optim_steps},  \ref{fig:layers} and \ref{fig:style_transfer} are created with their differentiable renderer.

\textbf{``Stylized neural painting” \cite{zou2021stylized}} use neural renderers (Figure \ref{fig:neural_painting}) of various stroke models (Figure \ref{fig:brushes_paintings}) to optimize stroke parameters. Optimal transport \cite{cuturi2013sinkhorn} and $L_1$ losses are used to guide their stroke updates. A style loss can be added to allow highly stylized paintings, according to a reference artwork. They use a special layered coarse-to-fine drawing approach that helps to cover the whole canvas with appropriately sized strokes.

\textbf{``From Objects to a Whole Painting” \cite{wang2021objects}} use a similar approach to \cite{zou2021stylized}, but try to draw one object after another through the use of object detection, which helps to eliminate artifacts and can be used to increase detail in important areas. Additionally, they add a dispersion loss, which helps to resolve gradient problems with overlapping strokes.

\textbf{``Rethinking style transfer: From pixels to parameterized brushstrokes” \cite{kotovenko2021rethinking}} specifically tailor their approach to solving artistic style transfer with brushstrokes. Instead of layering big and small strokes, they produce paintings made up of a large number of similar-sized lines, optimized through perceptual losses for style and content (Figure  \ref{fig:style_transfer}). They further enable control of stroke orientations through the use of user-supplied orientation lines.  

\textbf{``Differentiable drawing and sketching” \cite{mihai2021differentiable}} explore and compare the effect of using different losses, such as $L_2$ and LPIPS \cite{zhang2018unreasonable} to optimize stroke parameters through their differentiable rendering framework.

\textbf{``CLIPDraw: Exploring text-to-drawing synthesis through language-image encoders” \cite{frans2021clipdraw}} 
draw images according to specifications in text prompts, utilizing a CLIP loss and optimizing the parameters of differentiable strokes through gradient descent 
(Figures  \ref{fig:balloon_white_loss},  \ref{fig:balloon_loss},  \ref{fig:mona_lisa_loss}). They investigate the effect of different text inputs in combination with varying numbers of strokes and compare their algorithm with pixel-based approaches. \cite{schaldenbrand2022styleclipdraw} extend this work by adding a style loss to the optimization. This enables painting the text content in the style of a reference image.

\textbf{``CLIP-CLOP: CLIP-Guided Collage and Photomontage” \cite{mirowski2022clip}} use image patches from photos instead of strokes and optimize their color, position, and size through multiple CLIP critics. By limiting the field of view of critics with different text inputs, they are able to place objects at different positions in the painting. For example, one CLIP loss is applied with the input "Sky and sun" to the top left, and another one optimizes for "Landscape" at the bottom, forming a compound loss which results in a landscape painting with the sun in the top left corner.

\subsubsection{Deep learning painting algorithms}
\label{DL_painting}

\begin{figure}[!t]
\centering
\includegraphics[width=3.5in]{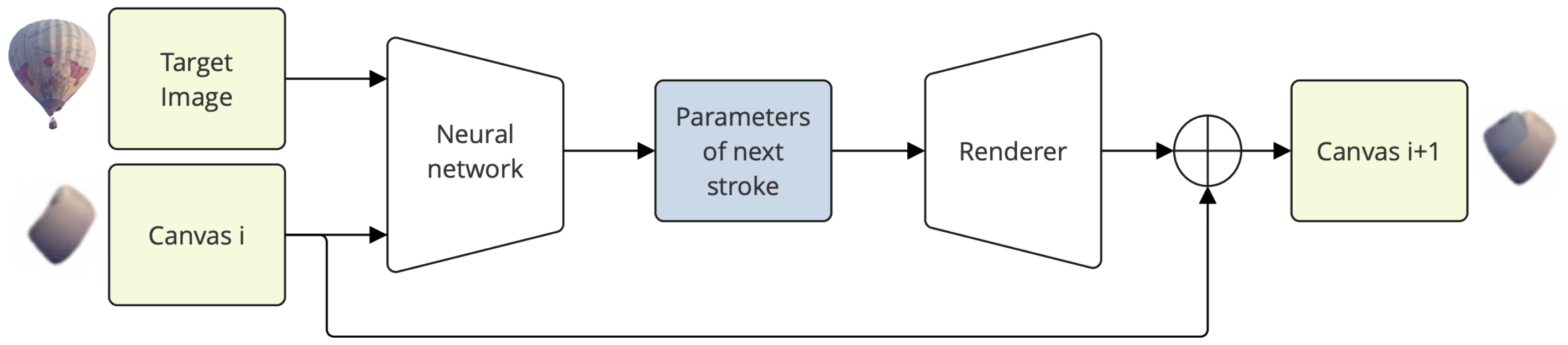}
\caption{During inference, deep learning algorithms usually predict the next stroke parameters based on the target image and the current canvas.}
\label{fig:deep_learning}
\end{figure}

Given a target image $I_T$, learning-based painting algorithms aim to find a mapping $P(I_T) = S$ such that the resulting rendered painting $I_P = R(S)$ has a small distance to the target: $\min dist(I_T, I_P)$. This distance is quantified, for example, through a loss function $L(S, I_T)$, which scores the predicted stroke parameters based on the target image. It is not clear how to directly calculate a loss between stroke parameters and pixel images, which is why the stroke parameters have to be rendered into a pixel image before being compared. After this, common image losses (section \ref{losses}) such as $L_2(S, I_T) = ||R(S) - I_T||_2$ can be used  to evaluate the quality of a painting.

While details may vary, the inference for most painting models is implemented very similarly (Figure \ref{fig:deep_learning}): given a target image $I_T$ and the current canvas $C_i$, they predict the next stroke parameters $S_{i+1}$to be rendered onto the canvas through a rendering engine. The new strokes are always painted on top of the older strokes by the rendering process, to generate a new canvas: $C_{i+1} = R(C_i, S_{i+1})$.
Training such a model is either done through supervised or reinforcement learning, with many different architectures, such as feedforward and recurrent neural networks, as well as transformers being used.

\smallskip

\textbf{Supervised learning}

\smallskip

\begin{figure}[!t]
\centering
\includegraphics[width=3.5in]{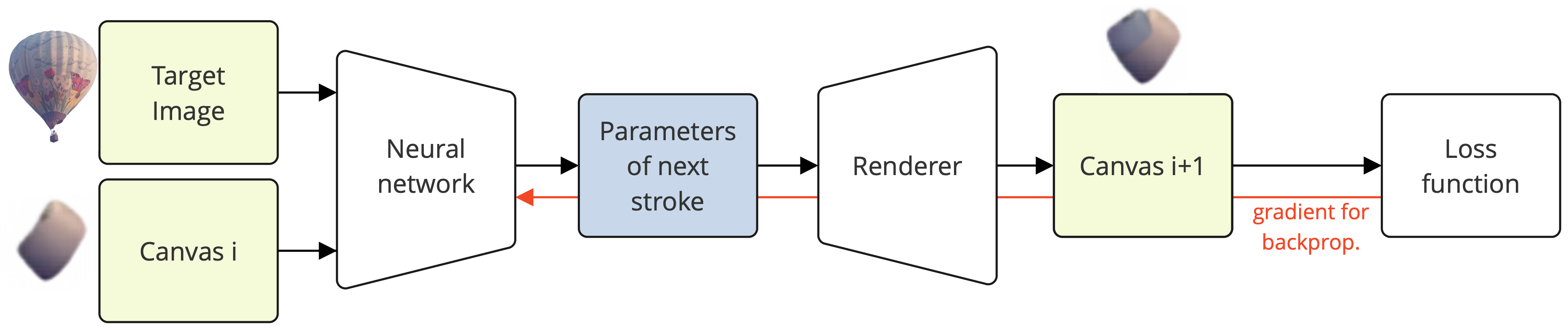}
\caption{Training procedure of a supervised learning model. The gradient of the loss function with respect to parameters of the neural network can be calculated through the use of differentiable rendering.}
\label{fig:supervised_learning}
\end{figure}

\begin{figure}[!t]
\centering
\includegraphics[width=3.5in]{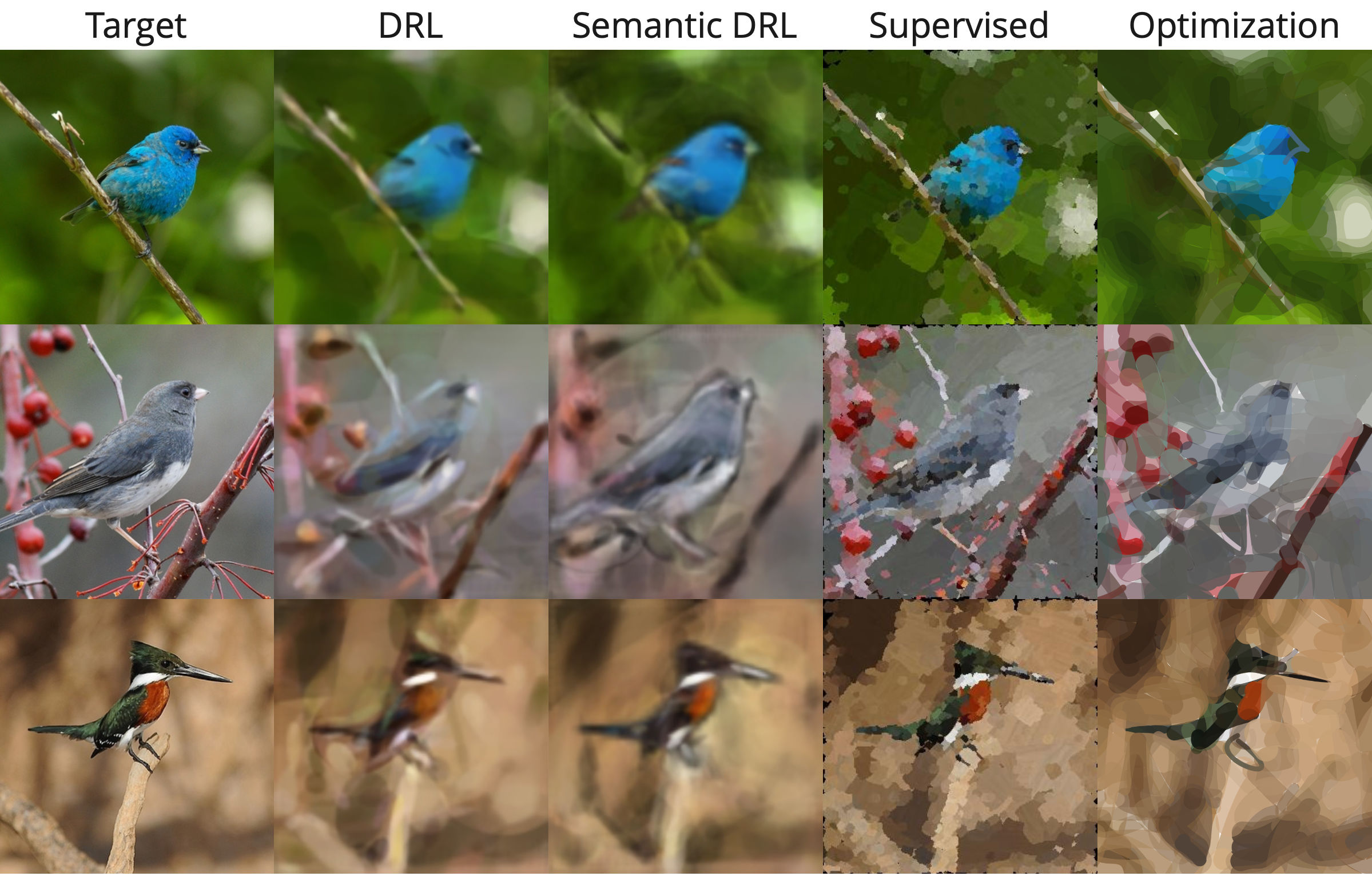}
\caption{Samples of the CUB-200 Birds dataset \cite{wah2011caltech}, painted with 150 strokes. The semantic DRL painter \cite{singh2021combining} is able to handle disadvantageous backgrounds better than the DRL baseline \cite{huang2019learning} and the supervised transformer model \cite{liu2021paint}. The optimization approach \cite{li2020differentiable} achieves good results, but is much slower than the trained agents.}
\label{fig:DL_comparison}
\end{figure}

In supervised deep learning, the weights of a neural network painter P are optimized by propagating a loss through the network. For backpropagation to work, the derivative $\frac{\partial L}{\partial s}$ of the loss function needs to be available during training (Figure \ref{fig:supervised_learning}). With a traditional pixel-based loss (e.g. $L= L_1(C_{i+1}, I_T) - L_1(C_i, I_T)$), this is only possible if the rendering engine $R(S)$ is differentiable (section \ref{differentiable_rendering}). In the case of simple sketch drawings, stroke-based training data $S^*$ is available and models can be trained through a stroke loss $L(S^*, \hat{S})$ 
without the need for any rendering engine. For example, \textit{sketch-rnn} \cite{ha2017neural} is a recurrent neural network trained to reconstruct simple doodles from their \textit{QuickDraw} dataset. Paired photo-sketch datasets are also available \cite{yu2017sketchx}, which enable supervised image-to-sketch translation \cite{song2018learning}. However, these are all constrained to simple sketch drawings and do not contain more complex paintings.

\textbf{``Unsupervised image to sequence translation with canvas-drawer networks” \cite{frans2018unsupervised}} train a recurrent neural network (RNN, called \textit{``Drawer"}) to predict a sequence of stroke parameters for a pixel image. The strokes are rendered through a differentiable rendering engine (\textit{``Canvas"}) and the final painting should minimize the $L_2$ loss to the target image. The model repeatedly predicts a new stroke to paint, based on the previously drawn canvas. To simplify the drawing process, the RNN only observes and draws a small region of the target image at once.

\textbf{``Strokenet: A neural painting environment” \cite{zheng2018strokenet}} train a recurrent agent to reconstruct images using curved strokes. The main contribution is a neural rendering architecture that supports complex strokes with many control points, instead of just two or three. The recurrent agent is trained to minimize the $L_2$ distance to a target image by using these long and smooth lines. Although their approach works well for simple images such as sketches and symbols, the recurrent drawing agent does not fully generalize to the drawing of natural images.

\textbf{``Paint transformer: Feed forward neural painting with stroke prediction” \cite{liu2021paint}} (Figure \ref{fig:DL_comparison}) employ a self-supervised painting approach, based on a transformer architecture. This enables near real-time prediction of strokes and stable training without any off-the-shelf dataset. During inference, the target image is split into layered patches for a coarse-to-fine painting procedure. The transformer predicts not just a set of stroke parameters for each image patch, but stroke confidences as well, which are used to decide which strokes are actually painted on the canvas. Training is achieved through a self-supervised pipeline, where the transformer recreates randomly generated stroke images, instead of real photos. The model can then be trained to optimize $L = L_1(I_T, I_P) + L_{Stroke}(S_T, S_P)$, because the ground truth strokes of the stoke-based target images are known. The model generalizes well to natural images, despite not being trained on them.

\smallskip

\textbf{Deep reinforcement learning}

\smallskip

\begin{figure}[!t]
\centering
\includegraphics[width=3.5in]{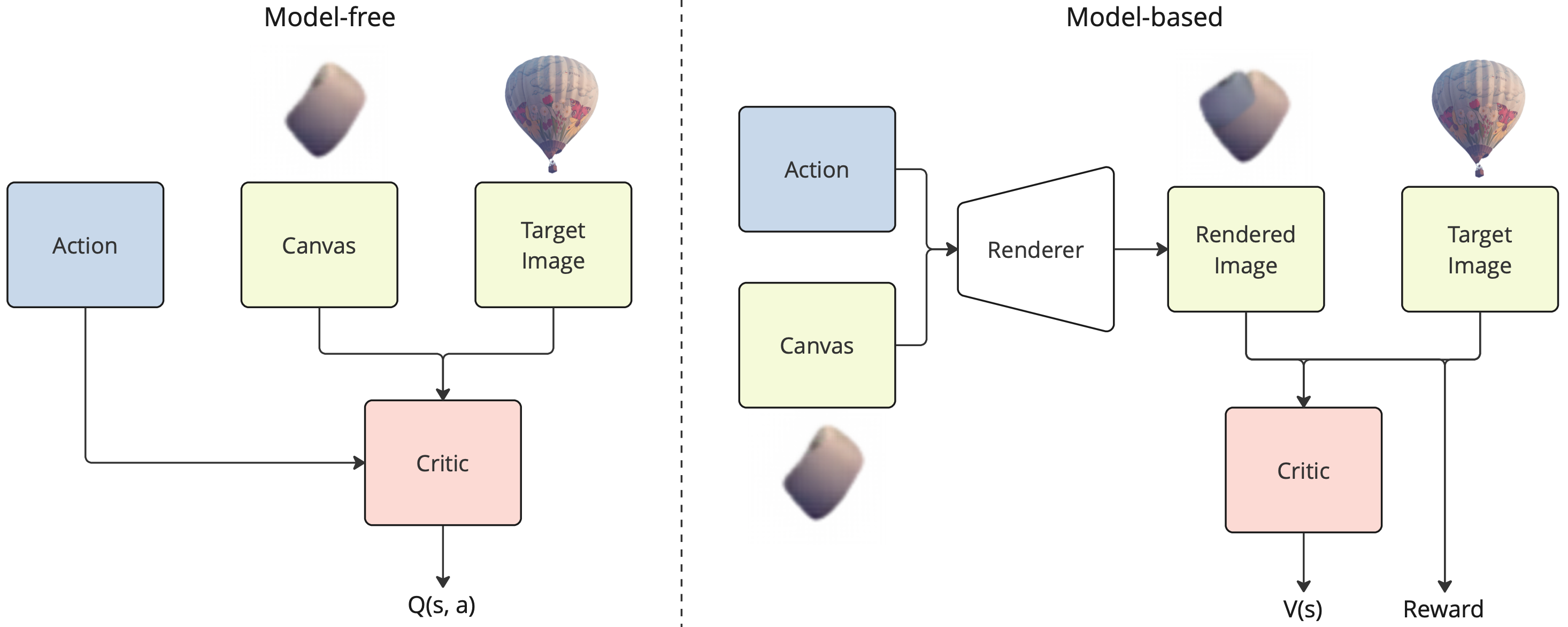}
\caption{Adapted from \cite{huang2019learning}: model-free and model-based DRL painting agents. In model-based DRL, the next action can be rendered onto the canvas before being judged by the critic, and the reward of an action can directly be calculated. This is enabled through differentiable rendering and results in more fine-grained reward signals.}
\label{fig:modelfree_vs_modelbased}
\end{figure}

In deep reinforcement learning (DRL) \cite{bach2020learn}, the painting task is modeled as a Markov Decision Process and does not always use differentiable rendering. The painting model $P$ is an agent that can observe the state $\mathbbm{s}_t = (C_t, I_T)$ of its environment and interacts with it through painting actions $\mathbbm{a}$ by predicting stroke parameters $S$ (Figure \ref{fig:deep_learning}). The rendering engine $R(S)$ implements the transition from one state to the next by executing the painting action $\mathbbm{a}$ and producing the new canvas $C_{t+1}$ by rendering the predicted stroke parameters $S_{t+1}$ onto the canvas $C_t$. The goal of the agent is to find a sequence of actions (i.e. stroke parameters) $(\mathbbm{a}_0, \mathbbm{a}_1, ..., \mathbbm{a}_n)$ which produce a high sum of rewards $\textrm{R} = \sum_{t=0}^n \gamma^t r(\mathbbm{a}_t, \mathbbm{s}_t)$. Rewards are calculated by using loss functions and they quantify whether the final canvas $C_n$ looks similar to the target image $I_T$. They sometimes make use of intermediate rewards $r(\mathbbm{a}_t, \mathbbm{s}_t) = L_t - L_{t+1}$ in addition to the final reward $r(\mathbbm{a}_n, \mathbbm{s}_n) = L(I_T, C_n)$. The exact implementation of the rendering procedure and the design of the state and reward representation varies between different implementations. Usually, the reward is encoded in a value function $V(s_t)$ that sums up the expected future rewards for a given state. The action space for almost all parameters of the painting models is continuous, which is why they are trained with various policy gradient methods \cite{sutton1999policy}. For non-differentiable rendering engines and model-free RL, the value function has to be approximated through a neural network in order to provide gradients for back-propagation. In model-based RL, a differentiable rendering engine is used in the value function, which provides more fine-grained reward signals and results in better sample efficiency and convergence \cite{huang2019learning}. Figure \ref{fig:modelfree_vs_modelbased} shows a comparison of model-free and model-based architectures.

\smallskip

\textbf{Learning to plan stroke sequences}

\smallskip
Unlike most pixel-based approaches, stroke-based paintings are created by deep learning models in a sequential manner. As such, the order of strokes and the sequential step-by-step creation of the images is a major concern in the design of painting agents. Two desirable properties of the stroke sequence are, that early strokes are sensibly placed with latter strokes in mind \cite{mellor2019unsupervised}, and that the painting procedure is natural and human-like \cite{singh2021intelli, schaldenbrand2021content, singh2021combining}. Apart from increasing the final image quality, well planned-out and natural drawing sequences can be used to create ``time-lapse" videos of the painting process \cite{singh2021intelli}.

\begin{figure}[!t]
\centering
\includegraphics[width=3in]{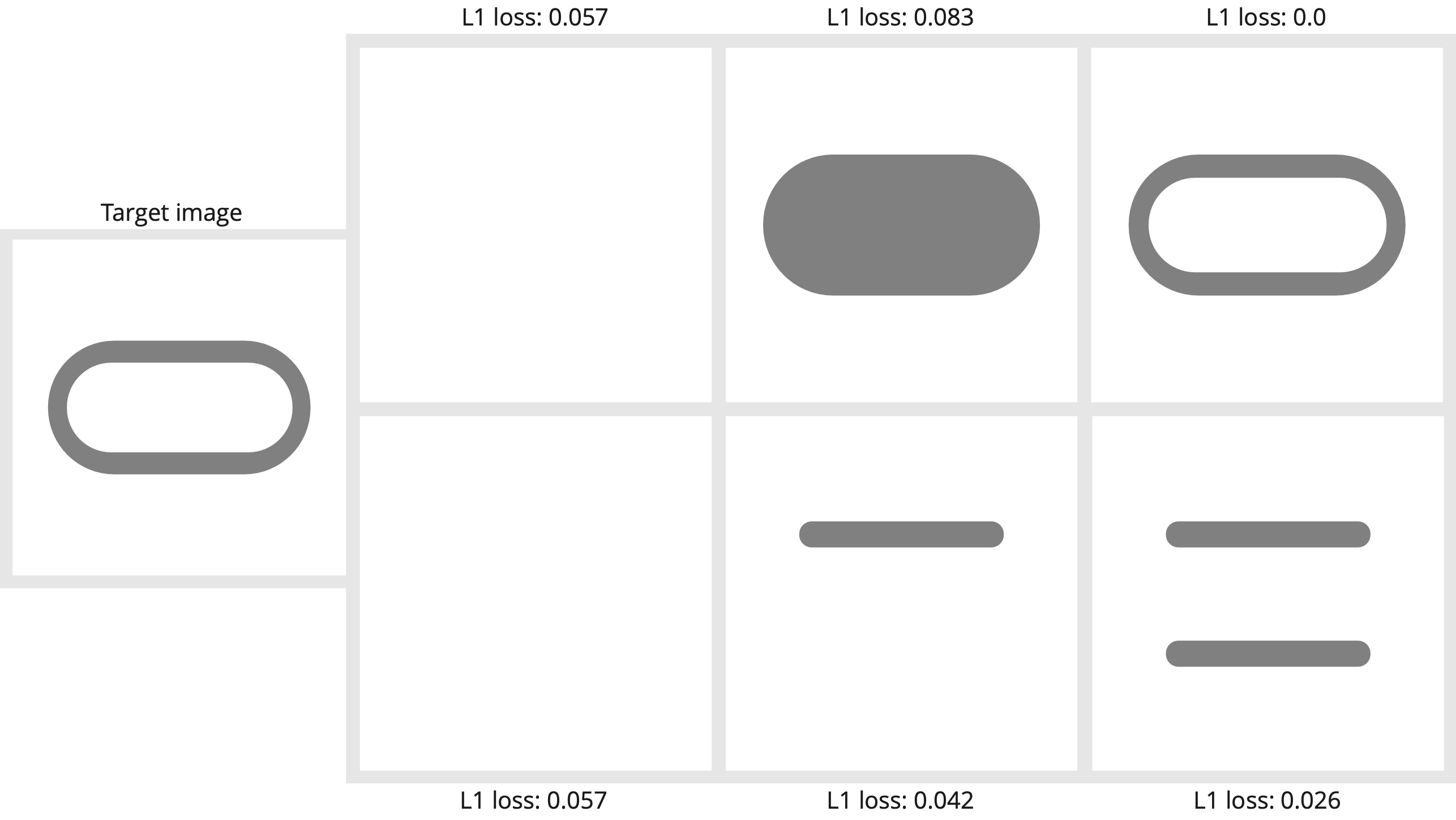}
\caption{Deep learning agents can learn painting sequences which will initially increase the image loss, but result in a good final drawing (top). Simply searching for a stroke that reduces the loss without regard for future strokes, might result in the bottom sequence. This local optimum is also a common result when using gradient descent for optimizing strokes.}
\label{fig:planning}
\end{figure}

\begin{figure}[!t]
\centering
\includegraphics[width=3.5in]{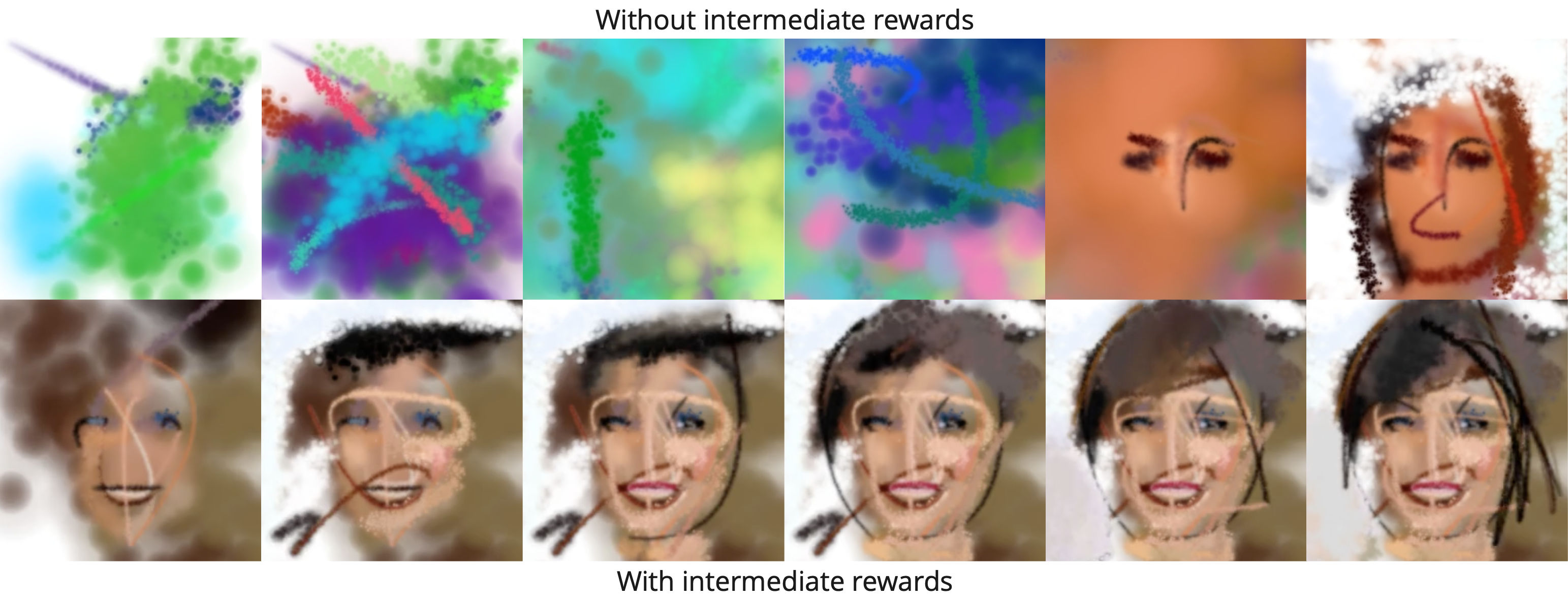}
\caption{Without intermediate rewards, agents struggle to make us of early strokes. With intermediate rewards and slightly more greedy painting policies, credit assignment with regard to earlier strokes is easier. Figures used with permission from \cite{mellor2019unsupervised}.}
\label{fig:intermediate_rewards}
\end{figure}

Optimization algorithms for painting are mostly designed to search for new stroke parameters that gradually reduce an error function. As such, they can get stuck in local optima  and struggle to put down strokes that would initially result in a higher loss, but produce a better painting in the long run (Figure \ref{fig:planning}). In deep learning and especially DRL, it is possible to train agents that are able to plan ahead \cite{beohar2022planning} and paint strokes which do not simply greedily reduce the loss function with each stroke, but will prioritize  a good final painting and regularly put down strokes that temporarily increase the loss \cite{mellor2019unsupervised} (Figure \ref{fig:planning}). DRL agents are trained to optimize the sum of discounted future rewards $\textrm{R} = \sum_{t=0}^n \gamma^t r(\mathbbm{a}_t, \mathbbm{s}_t)$. If the reward is only calculated based on the final painting $r(\mathbbm{a}_n, \mathbbm{s}_n) = L(I_T, C_n)$, the agent is able to learn non-greedy painting policies \cite{mellor2019unsupervised}. However, this hinders their ability to plan out long sequences of strokes due to the difficulty of assigning credit to earlier strokes. For example, \cite{mellor2019unsupervised} find that their agents only really learn to use the last 15 - 60 strokes to cover up the canvas and paint an image. Earlier strokes are completely random and hidden below the last strokes. This problem can be circumvented by introducing intermediate rewards $r(\mathbbm{a}_t, \mathbbm{s}_t) = L_t - L_{t+1}$ \cite{huang2019learning, mellor2019unsupervised} (Figure \ref{fig:intermediate_rewards}), which encourage believable earlier canvases at the cost of more greedy painting policies and allows agents to make use of over 1000 strokes. 

Another important factor that helps DRL agents to plan out their strokes is the state representation. Including a step number $T = \frac{t}{n} \in [0,1]$ in the state $\mathbbm{s}_t = (C_t, I_T, T)$ helps the agent to learn a natural painting progression with big strokes in the beginning and details in the end \cite{huang2019learning, xu2020demystify}. Without knowing about the current step number, agents struggle to add fine details to the painting. \cite{xu2020demystify}

A natural and human-like painting order can be designed to fulfill additional requirements. For example, agents can be trained to plan their stroke sequences with early recognizability of semantic objects in mind \cite{melnik2018world}, where the subject of the painting can be identified within a few strokes \cite{schaldenbrand2021content, singh2021combining}. Alternatively, the painting can be constructed from the background to the foreground, similar to how some humans paint. Here, the background is drawn first, and in the end, the main subject is added on top \cite{singh2021intelli}.

\smallskip

\textbf{Model-free DRL}

\smallskip

\begin{figure}[!t]
\centering
\includegraphics[width=3.5in]{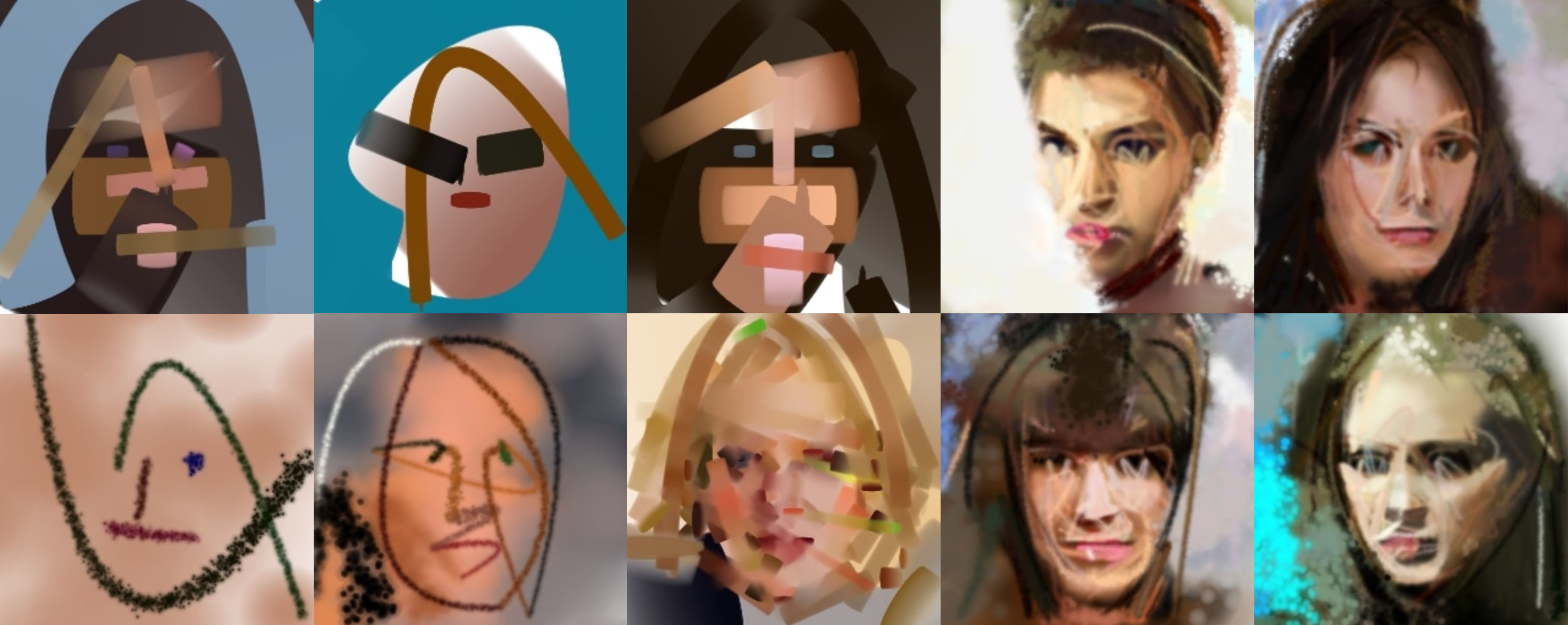}
\caption{The model-free painting agent from \cite{mellor2019unsupervised} is trained to create new unconditional samples for the Celeba-HQ dataset \cite{karras2017progressive}. Remarkably, the agent learns human-like visual abstractions, without ever having observed stick figures or stroke-based images in its training data. Figures used with permission from \cite{mellor2019unsupervised}.}
\label{fig:spiral_unconditional}
\end{figure}

\textbf{``Synthesizing programs for images using reinforced adversarial learning” (SPIRAL) \cite{ganin2018synthesizing}} use model-free deep reinforcement learning to train an agent capable of interacting with arbitrary non-differentiable painting programs. The agent's decision policy is implemented through an RNN $\pi_t = \pi(a_t|s_t)$ that predicts actions (e.g., next stroke parameters) based on the current state of the canvas. The value function $V^{\pi}(s_t)$ uses a WGAN loss to calculate the quality of the final canvas. The model is capable of interacting with arbitrary rendering programs and can be conditioned to reconstruct simple target images. Additionally, the WGAN loss enables unconditional image generation (similar to Figure \ref{fig:spiral_unconditional}). However, the agent is not able to faithfully recreate more complex and detailed images, such as portraits.

\begin{figure}[!t]
\centering
\includegraphics[width=3.5in]{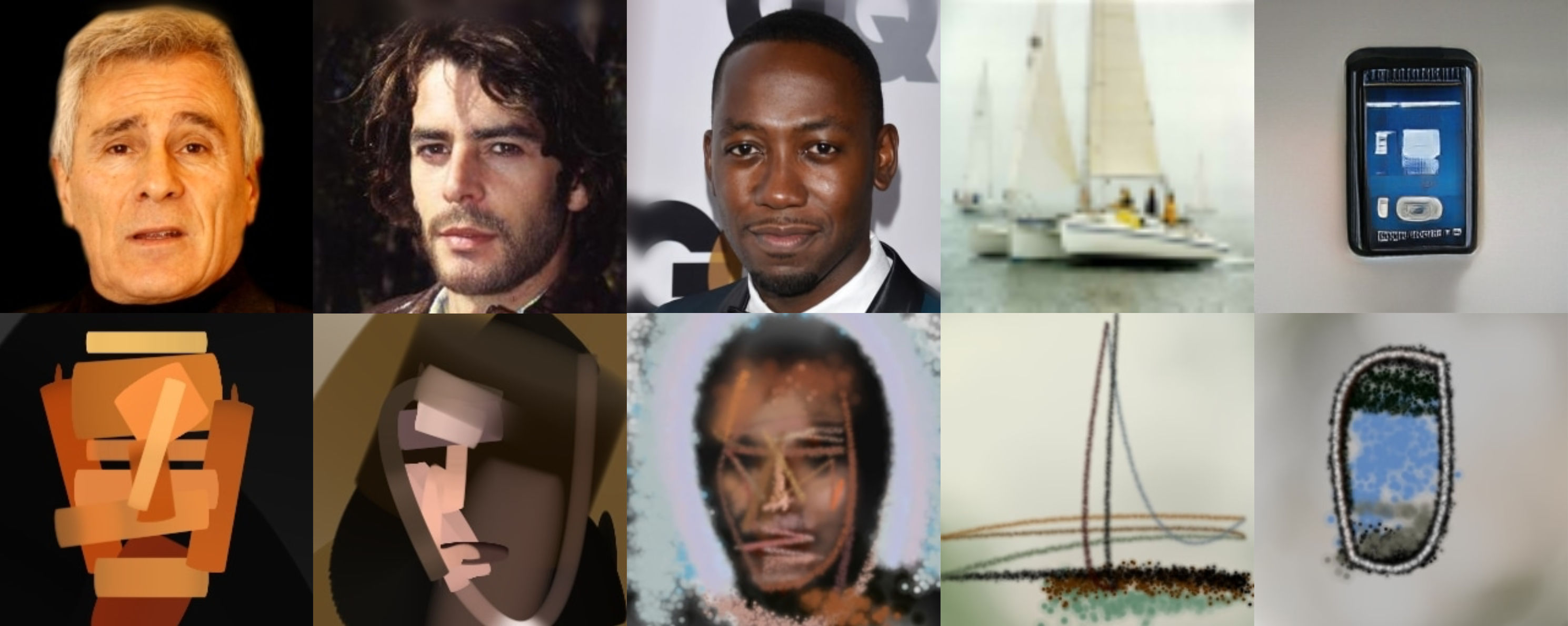}
\caption{The \textit{complement discriminator} from \cite{mellor2019unsupervised} allows their model-free painting agent to paint images with semantic similarity to the target images, instead of simple pixel similarity. Figures used with permission from \cite{mellor2019unsupervised}.}
\label{fig:spiral_conditional}
\end{figure}

\textbf{``Unsupervised doodling and painting with improved spiral” (SPIRAL++) \cite{mellor2019unsupervised}} (Figures \ref{fig:spiral_conditional}, \ref{fig:spiral_unconditional}) refine the previous architecture and achieve a model that can paint images with interesting visual abstractions and a high level of detail. They modify the discriminator architecture to rely much more on semantic similarity, instead of low-level pixel differences. This is achieved through carefully compositing the target image and painting before feeding them to the discriminator and results in a generator that draws bold and clear lines, instead of blurry shapes. Unlike most other DRL approaches, they are able to generate high-quality unconditional image samples (Figure \ref{fig:spiral_unconditional}) in addition to reconstructing target images (Figure \ref{fig:spiral_conditional}).

\textbf{``Paintbot: A reinforcement learning approach for natural media painting” \cite{jia2019paintbot}} train a painting agent that is able to paint arbitrarily sized natural images with a non-differentiable painting environment. They simplify the painting task by not feeding the agent the whole target and canvas to predict the next stroke, but only small image patches centered around the last brushstroke. The resolution of the canvas is only gradually increased to facilitate a coarse-to-fine painting process. The discounted rewards in the value function can be calculated through $L_2$, $L_{\frac{1}{2}}$ and perceptual losses. 

\textbf{``LpaintB: Learning to paint from self-supervision” \cite{jia2019lpaintb}} expand on their earlier work \cite{jia2019paintbot} through the use of self-supervised training. Ideally, the agent would be trained to mimic the actions of an expert painter. Because there is no paired dataset of images and paintings available, they use the attempts of an unsupervised RL agent as training data. For this, the target image to be painted is replaced with the final painting of the unsupervised agent. This solves the problem of sparse positive reward signals during the training of a painting agent by converting negative samples into positive ones. Overall, this combination of supervised and unsupervised agents in a feedback loop results in better performance and convergence.

\textbf{Model-based DRL}

\begin{figure}[!t]
\centering
\includegraphics[width=3.5in]{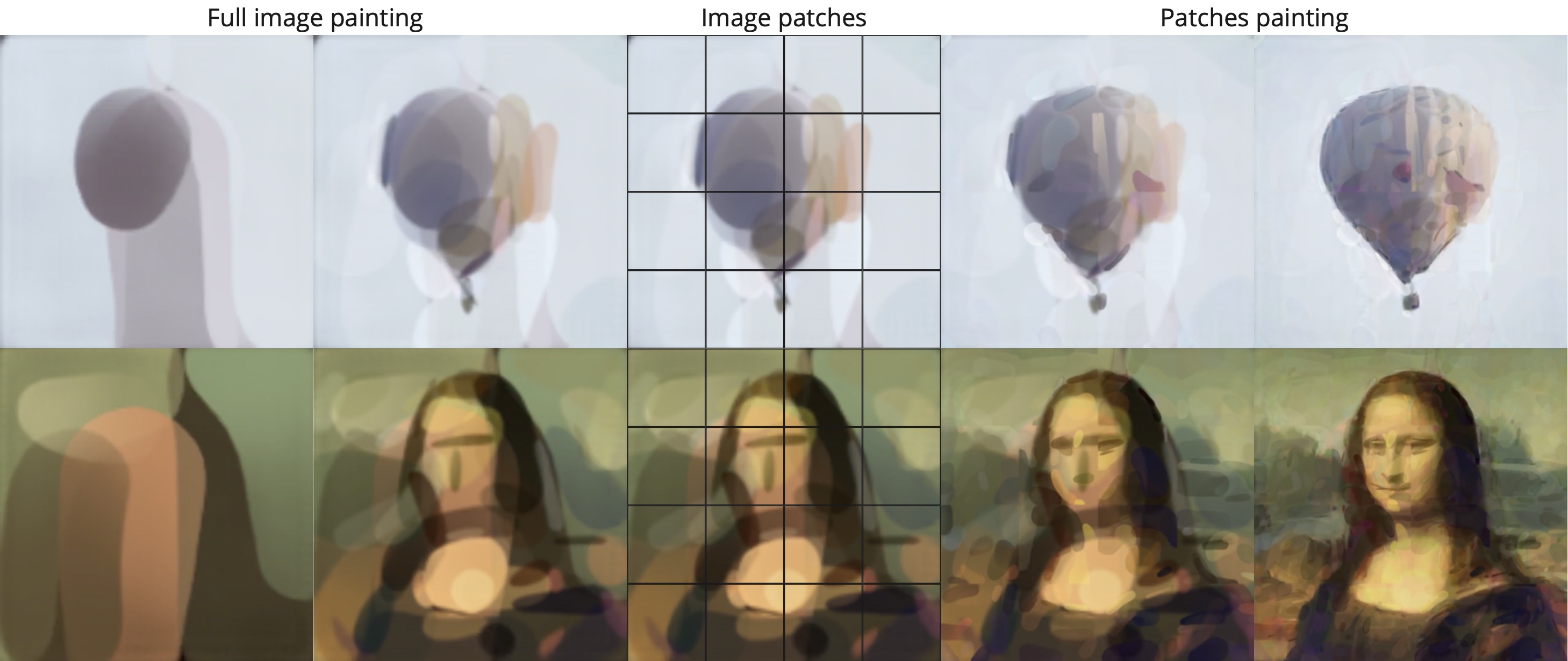}
\caption{Reinforcement learning agents paint an image by repeatedly placing strokes on a canvas. In order to simplify the painting process, the target image can be split up into small patches, which are individually painted. For example, \cite{huang2019learning} first paints some background strokes of the whole image (left) and then splits the images into patches (right) to paint smaller details. Top: 500 strokes; bottom: 1000 strokes.}
\label{fig:learn2paint}
\end{figure}

\textbf{``Learning to paint with model-based deep reinforcement learning” \cite{huang2019learning}} are the first DRL approach to make use of differentiable rendering for the reward calculation in their value function $V(s_t) = r(s_t,a_t) + \gamma V(s_{t+1})$ (see Figure \ref{fig:modelfree_vs_modelbased}). Through this, the reward $r$ is differentiable, and only the discounted value function for the next state is approximated through a neural network. This results in more fine-grained gradients, leading to better sample efficiency and convergence, compared to model-free RL. They achieve the best results with a WGAN loss, but $L_2$ works as well. The target image can be split into patches to simplify the painting process (see Figure \ref{fig:learn2paint}).

\textbf{``Demystify Painting with RL" \cite{xu2020demystify}} provide an overview and comparison of different hyperparameter choices for painting with model-based DRL. To achieve stable training, they find that deterministic policy gradients, predicting more than one stroke at once, and a state representation $s_t = (C_t, I, T)$ that includes a time-step $T$, are all important. For rewards, they experiment with $L_2$, $L_1$, WGAN, and perceptual style and content losses. They conclude that WGAN works best and that perceptual losses do not provide strong enough gradients for good results. Additionally, models trained on portraits will not necessarily generalize well to e.g. landscape images, even when the rewards are purely based on pixels.

\textbf{``Content masked loss: Human-like brush stroke planning in a reinforcement learning painting agent” \cite{schaldenbrand2021content}} experiment with different losses and rewards to make paintings more recognizable at the beginning of the drawing process. They extract semantically meaningful areas in the image through the layer activations from a pre-trained neural network and use them to weigh $L_1$ and $L_2$ losses. The models trained with these losses tend to draw important regions of the image first, similar to human painters.  

\textbf{``Combining semantic guidance and deep reinforcement learning for generating human level paintings” \cite{singh2021combining}} (Figure \ref{fig:DL_comparison}) propose a semantic guidance pipeline to improve the ability of model-based RL agents to understand foreground and background objects. The model receives separate rewards for foreground and background strokes, can efficiently attend to differently sized objects in the image, and learns to paint small important details of objects more carefully. This results in a human-like painting procedure with clearly defined and separated objects.

\textbf{``Intelli-Paint: Towards Developing Human-like Painting Agents” \cite{singh2021intelli}} build on their previous work \cite{singh2021combining} and build an agent with an even more content-driven painting strategy. Like a human painter, the agent starts painting the whole background by only concentrating on regions with low saliency. Afterwards, they paint one foreground object after another through a moving attention window. Finally, during inference, they remove and refine redundant strokes through direct stroke optimization. Overall, their agent is able to paint images with human-like stroke sequences and achieves good results with a much lower number of strokes, compared to previous approaches.

\subsection{Video painting algorithms}
\label{video_painting}

\begin{figure}[!t]
\centering
\includegraphics[width=3.5in]{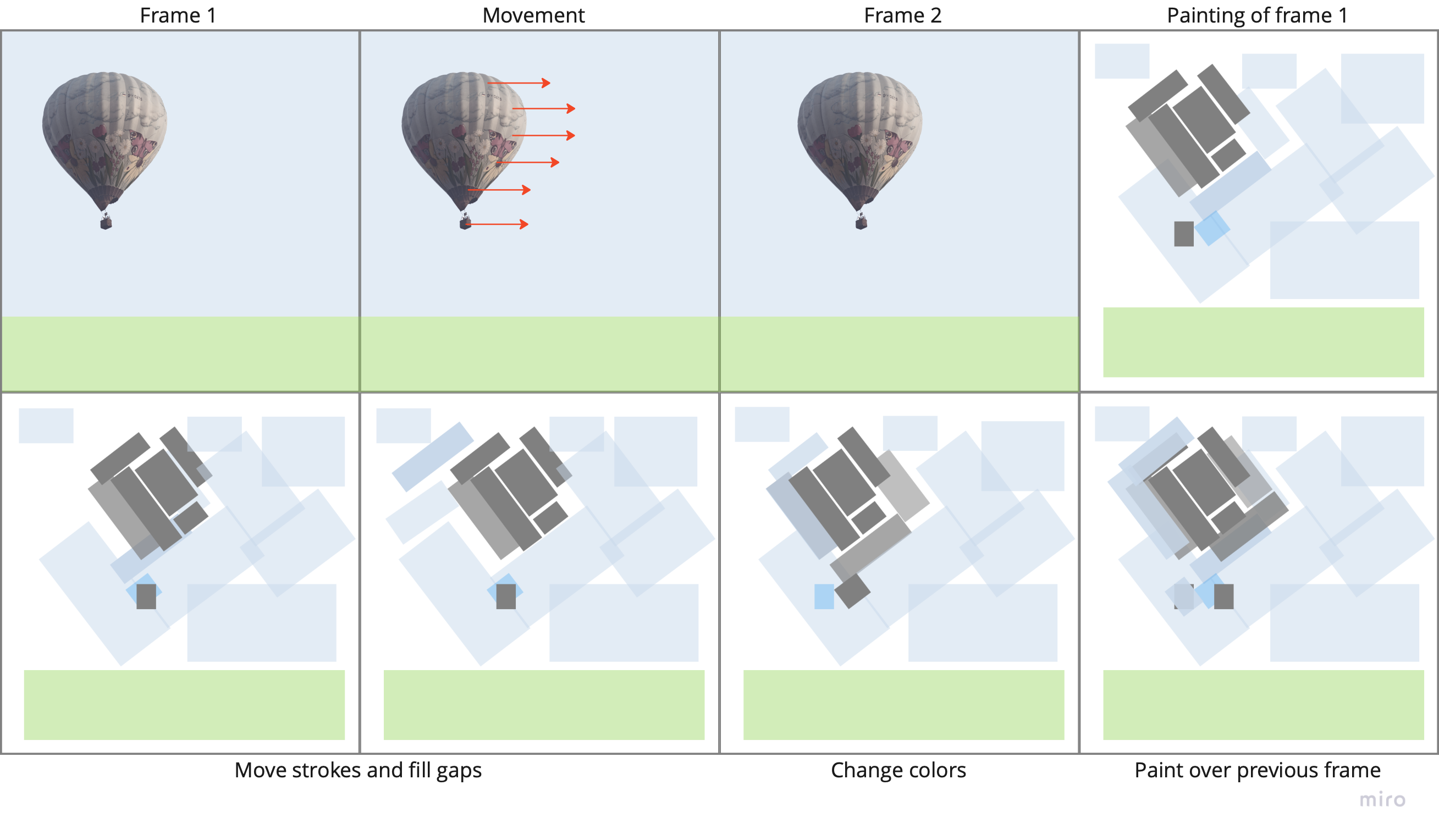}
\caption{After painting the first video frame, there are several possibilities for drawing the second frame. The goal is usually to have a video sequence without flickering and noise artifacts. 1: Strokes can be moved according to the optical flow in the target video, which can lead to overlapping strokes and gaps in the image. 2: The strokes are not moved, but only recolored to match the new frame. This results in a ``stained-glass" \cite{litwinowicz} effect. 3: New strokes are painted on top of the previous canvas in areas where the difference between canvas and target image is large.}
\label{fig:video_painting}
\end{figure}

One natural use case of stroke-based rendering is the automated painting of video sequences, which would be tedious and time-consuming for humans to do. In theory, any SBR algorithm could simply be used on individual video frames to produce a painted video. However, there are some attempts to reduce the resulting flickering \cite{hertzmann2000} by taking into account the similarities and movement between neighboring frames. The techniques are often based on optical flow, where the movement of pixels over time is estimated \cite{szeliski2010computer}. The idea is that strokes should follow the motion of objects within the image because otherwise, the rendering will have a "stained-glass" effect\cite{litwinowicz}. The following overview of algorithms is restricted to those features, which specifically support video painting. 
Figure \ref{fig:video_painting} shows some examples of commonly used video painting techniques. %

\textbf{``Processing images and video for an impressionist effect” \cite{litwinowicz}} displaces the stroke positions according to the optical flow of the video and removes those that are pushed out of the boundaries of the canvas. Moving the strokes creates areas with too high or low stroke density, which is resolved by removing and adding strokes through triangulation techniques. 

\textbf{``Painterly rendering for video and interaction” \cite{hertzmann2000}} uses a "painting-over" approach for videos by retaining the same canvas over the course of the video. After painting the first frame, the new strokes in the following frames are placed only in those areas of the canvas, where the source video significantly changes. In addition to this, strokes can be moved through optical flow.

\textbf{``Image and video based painterly animation” \cite{hays2004}} move strokes according to optical flow and gradually add and remove them to reduce flickering. Gaps in the image are filled with new strokes, which appear over time by increasing their opacity at each time step. In a similar manner, the strokes are gradually removed if they overlap each other or if they are pushed into regions where they no longer match the image. By adding custom motion information to still images, photos with moving strokes can be created.

\textbf{``Video painting with space-time-varying style parameters” \cite{kagaya2010video}} extend the algorithm of \cite{hays2004} by calculating an edge-based tensor field for stroke orientation, which is smoothed over time to generate coherent stroke orientations. Additionally, they use semantic video segmentation to enable image stylization over space and time.

\textbf{``Motion map generation for maintaining the temporal coherence of brush strokes” \cite{park2007motion}} achieve temporal coherence by adjusting only the orientation map for their strokes through optical flow and block-based motion detection. Areas with large movement are painted over, and flickering of the image is reduced through blending neighboring frames with a multi-exposure method.

\textbf{``Interactive painterly stylization of images, videos and 3D animations” \cite{lu2010interactive}} adjust stroke positions according to optical flow and add or delete strokes to retain a target stroke density. To reduce flickering artifacts, gradient and edge information is averaged over multiple frames, the size of strokes remains fixed, and orientation is only changed by small amounts each frame to fit the new gradients below the stroke. Stroke insertion and deletion happen by gradually fading them in and out.

\textbf{``Painterly animation using video semantics and feature correspondence” \cite{lin2010painterly} \cite{lin2012video}}  propagate a semantic segmentation map through the video through a space-time cutout algorithm. Motion is not calculated with optical flow for individual pixels but for image segments. This motion is then used to displace, rotate and resize strokes by moving their control points. The resulting gaps between strokes are filled by carefully placing new strokes in the background or by slightly warping neighboring strokes. Flickering and stroke displacement is damped through an iterative energy minimization, which holds neighboring strokes together and discourages fast movement.

\textbf{``Anipaint: Interactive painterly animation from video” \cite{Donovan2011anipaint}} calculate the optical flow with a rotoscoping approach where user-supplied region boundaries of objects are tracked through the video. Strokes are placed on the canvas according to the boundaries, and the control points are propagated through optical flow. Temporal coherence is accomplished by computing stroke parameters with an energy function that penalizes strokes of changing size and shape.

\medskip
Even if there is no full video to be painted, motion information can be used to enhance still images. For example in a landscape painting of a river, there might be an intent to show the flow of the river through stroke parameters. For this, \cite{lee2009motion} align the orientation of the brushstrokes in an image with the direction of motion. If there is no motion in an area, the strokes are aligned with the image gradients. \cite{hays2004} use optical flow to move brushstrokes while keeping the target image fixed. This creates an animated photo, where brushstrokes in selected areas are constantly moving, appearing, and disappearing. Motion information for these techniques can be either extracted from surrounding video frames or hand-crafted and supplied by the user.

\section{Conclusion}
The aim of this survey is to give a comprehensive overview of past and present approaches for generating painterly renderings through the use of computer vision, optimization, and deep learning. In addition to describing a wide range of different algorithms and organizing them in a taxonomy, explanations about common challenges are presented as well.

Stroke-based rendering algorithms are designed to generate images by calculating a sequence of strokes that can be rendered onto a digital canvas. This results in images akin to vector graphics which consist of parameterized shapes that can be rendered to pixel images at arbitrary resolutions. In comparison to other pixel-based approaches to image-making, this process is much more similar to how humans paint and draw. Because of this, the presented algorithms tend to produce images that are not necessarily photorealistic, but have a painterly appearance. There is great potential in using the methods not just to paint images for humans, but also to interactively support them in their digital painting tasks. In addition, automated painting algorithms can be used to efficiently create videos and animations with a painterly look.

With this survey, we hope to help artists and researchers in selecting the appropriate stroke-based painting algorithms for their needs. Furthermore, we hope to give inspiration for the development of new models that utilize not just the latest developments from deep learning, but also take older approaches into consideration. Reevaluating heuristics and optimization techniques while keeping recent developments surrounding differentiable rendering in mind might lead to the creation of new and interesting stroke-based algorithms.

For future work, this survey could be extended to cover other stroke-based drawing styles, such as sketching and pen-and-ink illustrations, which employ other machine learning architectures due to the existence of large stroke-based sketch datasets.

\ifCLASSOPTIONcompsoc
 \section*{Acknowledgments}
 The project has received funding from the KI-Starter initiative of the Ministry of Culture and Science of North Rhine-Westphalia.
\else
  regular IEEE prefers the singular form
 \section*{Acknowledgment}
\fi

\ifCLASSOPTIONcaptionsoff
  \newpage
\fi

\bibliographystyle{IEEEtran}
\bibliography{bibliography}

\end{document}